%% file: iclr2027_conference_v4.tex
\documentclass{article} 

\usepackage{needspace}
\usepackage{iclr2027_conference,times}

\usepackage{amsmath}
\usepackage{amssymb}
\usepackage{wasysym}
\usepackage{pifont}
\usepackage{bm}
\usepackage{longtable}
\usepackage{array}

\input{math_commands.tex}

\usepackage{graphicx}
\usepackage{booktabs}
\usepackage{tabularx}
\usepackage{multirow}
\usepackage{adjustbox}
\usepackage{float}
\usepackage{caption}
\usepackage{makecell}
\usepackage{soul}
\renewcommand{\hl}[1]{#1}
\sethlcolor{yellow!25}

\usepackage[table]{xcolor}
\usepackage{colortbl}

\definecolor{GroupGray}{HTML}{F2F2F5}
\definecolor{BestRed}{HTML}{B42318}
\definecolor{SecondPurple}{HTML}{5B3F99}

\newcommand{\best}[1]{\textbf{#1}}
\newcommand{\secondbest}[1]{\underline{#1}}

\usepackage{hyperref}
\usepackage{url}
\hypersetup{hidelinks}
\soulregister\citep7
\title{MetaRigor: From Agentic Automation to Verifiable End-to-End Meta-Analysis Through Scientific Decision Boundaries}

\makeatletter
\newcommand{\affilthanks}[2]{%
  \begingroup
  \def\@makefnmark{\textsuperscript{#1,\@thefnmark}}%
  \thanks{#2}%
  \endgroup
}
\makeatother

\author{%
  {\bfseries
    Yifeng Xie\affilthanks{1}{Equal contribution.},
    Chenke Li\textsuperscript{1,*},
    Yangshuo Miao\textsuperscript{1},
    Jingyuan Zheng\textsuperscript{1},
    Mingxuan Liu\textsuperscript{2},%
  }\\
  {\bfseries
    Feiwei Qin\textsuperscript{1,\textdagger},
    Yaqi Wang\textsuperscript{1},
    Huiyu Zhou\textsuperscript{3},
    Yifei Chen\affilthanks{2}{Corresponding authors: Yifei Chen (justlfc03@gmail.com) and Feiwei Qin (qinfeiwei@hdu.edu.cn).}%
  }\\
  {\normalfont
    \textsuperscript{1}Hangzhou Dianzi University\quad
    \textsuperscript{2}Tsinghua University\quad
    \textsuperscript{3}University of Leicester%
  }%
}

\iclrfinalcopy 

\begin{document}

\maketitle

\fancyhead{} 
\renewcommand{\headrulewidth}{0pt}

\begin{abstract}

Automated meta-analysis requires not only accurate numerical extraction but also the consistent preservation of result identity across stages, including the associated study, comparison, population, outcome, time point, and analysis conditions. We present MetaRigor, an LLM workflow framework for verifiable end-to-end evidence synthesis. MetaRigor organizes study results as Study $\times$ Target units and operationalizes scientific decision boundaries at the task level through Task Contracts, programmatic rules, and source binding. The model performs semantic interpretation that depends on study context, whereas programs execute formalizable rules, computations, validation, and source verification. We conducted controlled evaluations across five stages: evidence acquisition, data extraction, risk of bias assessment, certainty of evidence assessment, and manuscript generation. MetaRigor achieved an $F_1$ score of 86.29\% for complete result delivery and 80.20\% agreement for domain-level certainty of evidence judgments. Generated manuscripts received scores of 86.00 under human evaluation, exceeding all three baselines. \hl{A seven-case downstream end-to-end evaluation further yielded an aggregate $F_1$ score of 77.01\%, exceeding the strongest baseline by 9.78 percentage points}. Cross-system tracing of archived outputs from three systems further showed that local mismatches or omissions in result identity could alter statistical inputs and the objects subjected to risk of bias assessment, with these errors subsequently propagating into the final report. Ablation experiments showed that task constraints and programmatic rules played distinct roles across different stages. These findings indicate that, in LLM-driven evidence synthesis workflows, consistently maintaining explicit result identity, decision boundaries, and provenance relationships is critical to improving end-to-end verifiability. Source code is available at \url{https://anonymous.4open.science/r/Meta-Rigor-8C62}.
\end{abstract}

\begin{figure}[ht]
    \centering
    \includegraphics[width=\linewidth]{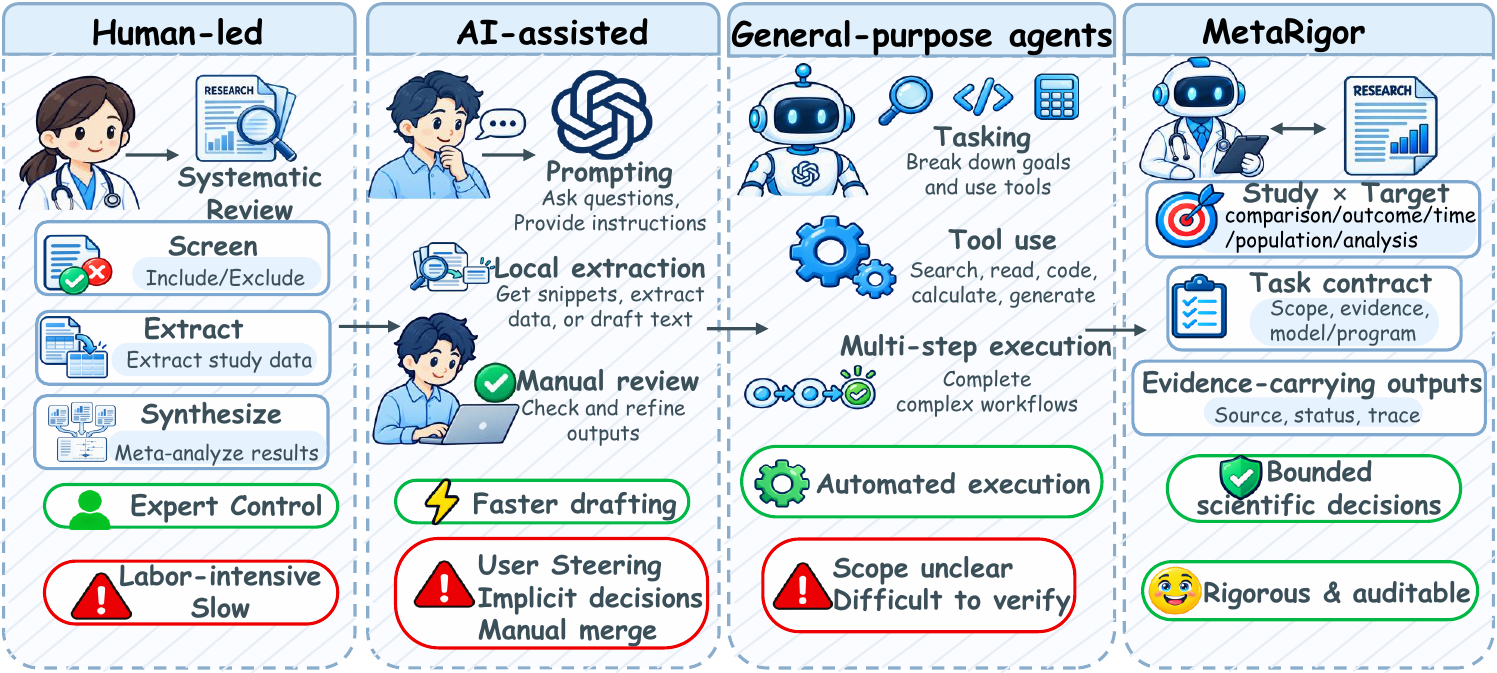}
    \caption{Evolution of meta-analysis paradigms. As meta-analysis evolves from human-led workflows to AI-assisted and general-purpose agents, automation increases while scientific decision boundaries may remain implicit. MetaRigor makes study identity, task constraints, and evidence provenance explicit to support rigorous and verifiable end-to-end evidence synthesis.}
    \label{fig:overview}
\end{figure}
\section{Introduction}

Meta-analysis synthesizes quantitatively comparable results from independent studies in a systematic review to estimate an overall effect and characterize heterogeneity and uncertainty across studies. A complete workflow extends beyond statistical synthesis to include interdependent stages: research question formulation, literature retrieval and screening, data extraction, risk of bias assessment, certainty of evidence assessment, and result reporting \citep{higgins2024cochrane,page2021prisma}. The Cochrane Handbook and PRISMA 2020 identify study identification, screening, appraisal, and synthesis as core systematic review components. Recently, large language models have accelerated automation across multiple evidence synthesis stages. Manalyzer coordinates multiple agents to organize analysis \citep{xu2025manalyzer}. meta-pipe explores full-workflow automation for systematic reviews and meta-analyses through an LLM-agent pipeline \citep{lin2026meta}. AutoSynthesis integrates structured retrieval, result extraction, and statistical synthesis \citep{taherinezhad2026autosynthesis}. Figure \ref{fig:overview} illustrates the progression from human-led workflows to AI-assisted systems and general-purpose agents. As automation advances, a key question is how a system can ensure that each stage operates on the correct study object, uses only task-admissible facts, and follows the methodological rules governing each decision while maintaining consistency across all stages.

A central challenge is that reports, studies, and target results do not map uniquely onto one another. Information from a single study may be distributed across the primary report, supplementary materials, trial registry records, and subsequent publications. Consequently, deduplicating reports alone cannot ensure that reports are correctly linked to the same underlying study, as discussed in \citep{li2024collecting}. The Cochrane Handbook explicitly treats the study, rather than the report, as the fundamental unit of a systematic review and requires multiple reports from the same study to be appropriately linked. Moreover, the same outcome may be reported under different comparisons, populations, time points, and analysis conditions. Thus, extracting the correct numerical value does not necessarily mean that the result required by the Protocol has been identified, as emphasized in \citep{mckenzie2024defining}. Relevant methodological guidance requires prespecified comparisons, outcomes, and time points, together with procedures for selecting among multiple eligible results. Once result identity is misaligned, downstream statistical synthesis, methodological assessment, and interpretation may diverge from the original research question.

\hl{We collectively define the Scientific Decision Boundary in terms of three elements: the object of judgment, the admissible facts, and the decision authority assigned to models and programs. This concept defines the constraints that govern scientific judgments at each stage.} For example, determining whether a comparison, study population, or time point matches a prespecified target requires the model to interpret the study context. By contrast, programs handle candidate enumeration, numerical computation, source verification, and execution of methodological rules that can be formalized. These constraints must be maintained in downstream stages to prevent generation from altering established result identities or overriding outputs produced by programmatic rules.

Guided by this principle, we propose MetaRigor, which operationalizes the Scientific Decision Boundary through structured representations and execution constraints within a unified framework. The system begins with a Protocol confirmed by the researcher. It consolidates the primary report, supplementary materials, trial registry records, and subsequent reports associated with the same study into a Study Packet. Each target result is then represented as a Study $\times$ Target pair, where Study denotes a specific study and Target is jointly defined by the comparison, population, outcome, time point, and analysis conditions. A Task Contract specifies the object of judgment, admissible facts, and delivery requirements for each task, while assigning distinct responsibilities to models and programs. MetaRigor also links each output to a retrievable location in the original source, thereby preserving the relationships among results, judgments, and supporting evidence and enabling their verification throughout the five stages: evidence acquisition, data extraction, risk of bias assessment, certainty of evidence assessment, and manuscript generation.

We evaluate MetaRigor through controlled evaluation of individual stages and analysis of propagation across stages. With upstream inputs held fixed, the controlled evaluation separately examines the recovery of study materials, result delivery, methodological judgments, and manuscript quality. Ablation experiments further examine the specific contributions of key constraints at different stages, while source verifiability is assessed independently. We also analyze continuous pipeline execution and archived artifacts from three systems to determine whether result identities and the states of unresolved items, issues, and missing information are preserved across stages. We then trace how local mismatches or omissions propagate to downstream statistical inputs, the targets of risk of bias assessment, and the final report.

The main contributions of this work are as follows:
\begin{itemize}
\item We introduce a result representation centered on Study $\times$ Target that preserves result identity and its defining conditions throughout material organization, result selection, methodological assessment, and manuscript generation.
\item We operationalize the Scientific Decision Boundary through task constraints and execution mechanisms that explicitly allocate responsibilities between models and programs, while supporting evidence verification across stages through source binding.
\item We establish a controlled evaluation framework covering five stages and conduct an analysis of artifacts from the complete workflow. Comparative and ablation experiments quantify the contributions of key constraints, while propagation analysis reveals how deviations in result identity affect downstream evidence synthesis.
\end{itemize}

\section{Related Work}

\textbf{Task-Specific Evidence Synthesis Methods.}
Early research on automated evidence synthesis primarily focused on
individual tasks within the workflow.
ASReview uses active learning to improve literature screening
efficiency \citep{van2021open}, while Trialstreamer automatically
identifies clinical trials and extracts structured information such
as PICO elements \citep{marshall2020trialstreamer}.
TrialMind and LEADS further leverage LLMs for literature retrieval,
eligibility assessment, and structured data extraction
\citep{wang2025accelerating,wang2025foundation}.
\hl{
These methods address individual components of evidence synthesis
rather than organizing an entire review protocol.
For example, ROBOTO2 uses supporting evidence from trial reports
to answer RoB 2 signaling questions and produce domain-level
assessments \citep{hevia2025roboto2}, while EviSearch extracts
structured information from trial papers and links extracted
fields to page-level evidence through attribution and verification
against the original text \citep{ahuja2026evisearch}.
The main distinction of MetaRigor lies in the scope of the problem
it addresses. A systematic review protocol may involve multiple
targets, each involving multiple included studies, while each
study may be represented by multiple reports.
In contrast, MetaRigor does not take an isolated task or an
individual report as its overall processing boundary.
Instead, it operates at the level of a complete systematic review
protocol, organizing multiple targets, studies, and their
associated reports within a shared structured research representation,
and maintaining their relationships across
all five stages: evidence acquisition, data extraction,
risk of bias assessment, evidence certainty assessment,
and report generation.
}

\textbf{General-Purpose Agent Frameworks. } In recent years, general agent frameworks have expanded the capabilities of large language models to handle complex tasks through tool use, code execution, and multi step task orchestration \citep{yao2022react,wang2025openhands}. ReAct interleaves reasoning and action, allowing models to adapt subsequent steps based on tool execution results. OpenHands provides a general agent execution framework that supports code editing, command execution, and interaction with the execution environment. Codex CLI and OpenHands CLI are representative general purpose agent environments that leverage external tools and code execution to complete complex tasks. These frameworks primarily provide general task execution and tool interaction capabilities. Their specific task objectives, domain knowledge, and evaluation criteria typically depend on prompts and external tools provided by users. They are not specifically designed for the methodological workflows of systematic reviews and meta-analyses.

\hl{\textbf{Automated Meta-Analysis Systems Covering the Full Workflow. } Recent work on LLM agents has begun extending from individual tasks to full workflow automation of meta-analysis. Table~\ref{tab:system_coverage} summarizes the functional coverage and code availability of these systems and tools. Manalyzer integrates literature retrieval, screening, data extraction, and report generation, but its default statistical analysis module lacks a standard procedure for effect size pooling \citep{xu2025manalyzer}. AutoSynthesis further supports effect size calculation, random effects pooling, and risk of bias assessment, but lacks a dedicated GRADE assessment stage \citep{taherinezhad2026autosynthesis}. Meta-pipe adopts a staged Skill workflow orchestrated by Claude Code, covering literature retrieval, screening, extraction, statistical synthesis, and manuscript generation, but does not report formal performance validation \citep{lin2026meta}. Despite these advances, comprehensive evaluations across all workflow stages using multiple published systematic reviews, and joint assessment of stage-level correctness and cross-stage consistency, remain lacking. To address this gap, MetaRigor establishes a benchmark based on 102 published systematic reviews, spanning five stages and covering 717 study inclusions. It evaluates individual stage performance and the consistency of study results across stages, with source-linked outputs supporting evidence verification and error tracing.}
\begin{table}[htbp]
\centering
\caption[Functional coverage and code availability of evidence synthesis systems and tools.]{%
\protect\hl{Functional coverage and code availability of evidence synthesis systems and tools.}%
}
\label{tab:system_coverage}

\begingroup
\normalsize
\setlength{\tabcolsep}{5pt}
\renewcommand{\arraystretch}{1.18}

\definecolor{StripeGray}{gray}{0.96}
\definecolor{SystemBlue}{HTML}{E3F0FA}
\definecolor{OursBlue}{HTML}{CCE3F5}

\rowcolors{2}{StripeGray}{white}

\begin{tabularx}{\linewidth}{
  @{\hspace{6pt}}
  l
  *{6}{>{\centering\arraybackslash}X}
  @{\hspace{6pt}}
}
\toprule
\rowcolor{white}
\textbf{System}
& \textbf{EA}
& \textbf{DE}
& \textbf{RoB}
& \textbf{EC}
& \textbf{MS}
& \textbf{Code} \\
\midrule

ASReview \citep{van2021open}
& $\checkmark$
& ---
& ---
& ---
& ---
& $\checkmark$ \\

Trialstreamer \citep{marshall2020trialstreamer}
& $\checkmark$
& \LEFTcircle
& ---
& ---
& ---
& $\checkmark$ \\

TrialMind \citep{wang2025accelerating}
& $\checkmark$
& $\checkmark$
& ---
& ---
& ---
& $\checkmark$ \\

LEADS \citep{wang2025foundation}
& $\checkmark$
& $\checkmark$
& ---
& ---
& ---
& $\checkmark$ \\

RobotReviewer \citep{marshall2016robotreviewer}
& \LEFTcircle
& \LEFTcircle
& $\checkmark$
& ---
& ---
& $\checkmark$ \\

ROBOTO2 \citep{hevia2025roboto2}
& ---
& \LEFTcircle
& $\checkmark$
& ---
& \LEFTcircle
& $\checkmark$ \\

EviSearch \citep{ahuja2026evisearch}
& ---
& $\checkmark$
& ---
& ---
& \LEFTcircle
& $\checkmark$ \\

\rowcolor{SystemBlue}
meta-pipe \citep{lin2026meta}
& $\checkmark$
& $\checkmark$
& $\checkmark$
& $\checkmark$
& $\checkmark$
& $\checkmark$ \\

\rowcolor{SystemBlue}
AutoSynthesis \citep{taherinezhad2026autosynthesis}
& $\checkmark$
& $\checkmark$
& $\checkmark$
& ---
& $\checkmark$
& \ding{55} \\

\rowcolor{SystemBlue}
Manalyzer \citep{xu2025manalyzer}
& $\checkmark$
& $\checkmark$
& ---
& \LEFTcircle
& $\checkmark$
& $\checkmark$ \\

\rowcolor{OursBlue}
\textbf{MetaRigor}
& $\checkmark$
& $\checkmark$
& $\checkmark$
& $\checkmark$
& $\checkmark$
& $\checkmark$ \\

\bottomrule
\end{tabularx}

\par\vspace{5pt}
\begin{minipage}{\linewidth}
\footnotesize
$\checkmark$: implemented or publicly available;
\LEFTcircle: partial coverage;
---: capability not confirmed;
\ding{55}: code not publicly available.
Code availability reflects the audit date.
\par\vspace{2pt}
\end{minipage}

\endgroup
\end{table}



\begin{figure}[!t]
    \centering
    \includegraphics[width=\linewidth]{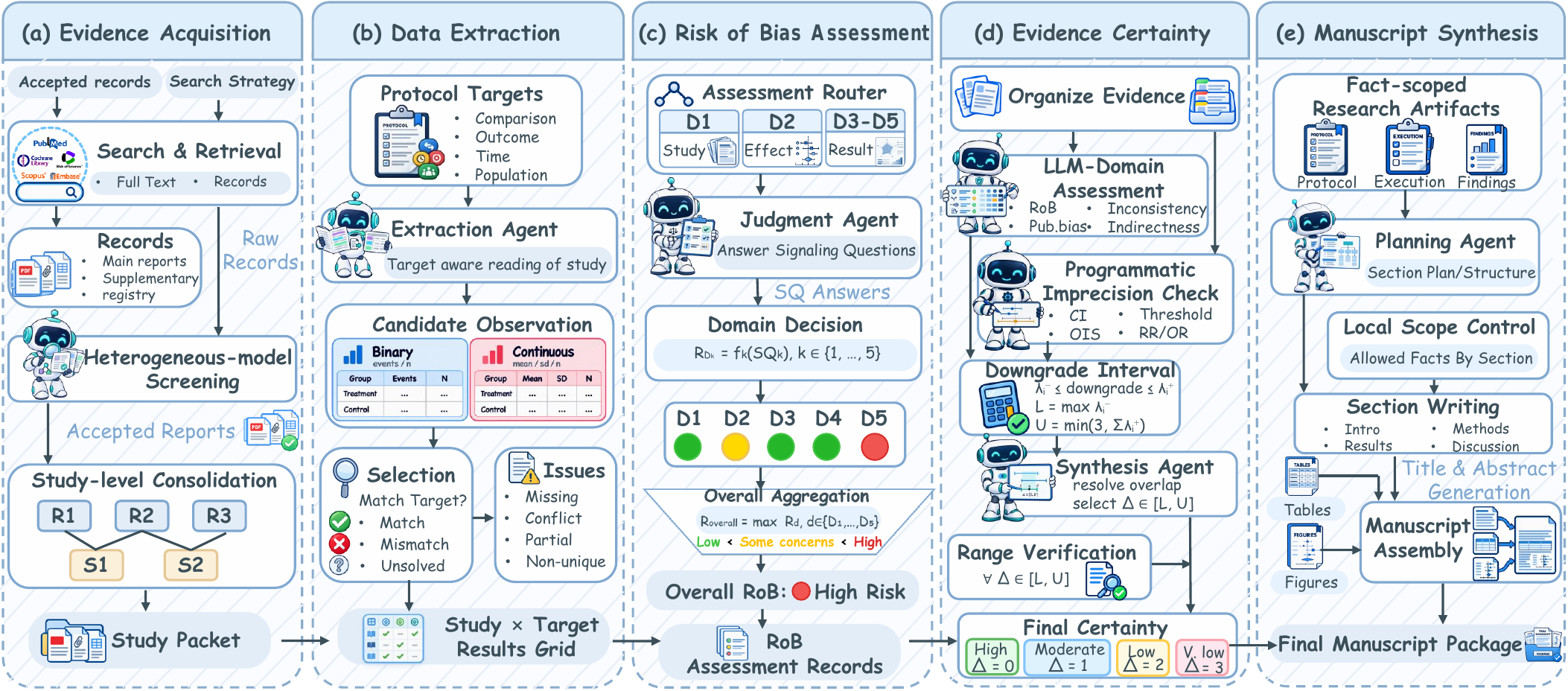}
    \caption{The complete evidence synthesis workflow of MetaRigor. (a) Evidence acquisition retrieves and consolidates relevant reports. (b) Data extraction identifies results for each Study $\times$ Target pair. (c) Risk of bias assessment evaluates the risk of bias in studies. (d) Certainty of evidence assessment evaluates the certainty of evidence. (e) Manuscript generation organizes research artifacts and generates the manuscript.
    }
    \label{fig:architecture}
\end{figure}


\section{Method}

As illustrated in Figure \ref{fig:architecture}, MetaRigor takes a Protocol confirmed by the researcher as its starting point and organizes the evidence synthesis workflow around study identity and result identity. The system first uses a shared research representation to unify the study objects and result indices passed across stages. It then applies the Scientific Decision Boundary to constrain judgments and execution at each stage, completing five stages: evidence acquisition, data extraction, risk of bias assessment, certainty of evidence assessment, and manuscript generation.
\subsection{Shared Research Representation}

MetaRigor uses the study as the unit for organizing materials. Given a study set $\mathcal{S}$ and a target set $\mathcal{T}$, the primary report, supplementary materials, registry records, and subsequent reports associated with the same study are consolidated into a Study Packet. Reports and studies may have many-to-many relationships, and the system preserves report identities and their associated links. Studies whose identities remain unresolved or whose materials are insufficient retain an unresolved status rather than being forcibly consolidated.

On the basis of this organization, MetaRigor identifies each result through a Study $\times$ Target pair. A Target represents a set of prespecified result conditions, denoted by $t=(\kappa,p,o,\tau,a)$, corresponding to the comparison, population, outcome, time point, and analytical conditions, respectively. The Cartesian product $\mathcal{G}=\mathcal{S}\times\mathcal{T}$ defines the index space of results to be analyzed. This representation does not assume that complete observations exist for every combination, but allows for partial results, missing results, and ambiguity.

Study $\times$ Target distinguishes prespecified analysis targets from observations actually reported in the sources. The system matches subsequently extracted Observations to their corresponding Study $\times$ Target pairs, thereby distinguishing candidate results within the same study that differ in comparison, outcome, time point, or analytical conditions. This provides a unified result index for data extraction, methodological assessment, and statistical synthesis.

\subsection{Operationalizing Scientific Decision Boundaries}

Building on the shared research representation, MetaRigor uses the Scientific Decision Boundary (SDB) to constrain judgments and execution at each stage. The SDB has three elements: the object of judgment, admissible facts, and decision authority assigned to models and programs. For example, determining whether a comparison, study population, or time point matches a target requires models to interpret the study context. By contrast, candidate enumeration, numerical computation, source verification, and methodological rules that can be formalized are executed by programs.

MetaRigor operationalizes the SDB for individual tasks through Task Contracts. Each contract specifies the object of judgment, admissible facts, the respective decision authority of models and programs, delivery requirements, and validation rules. Models handle factual interpretation and semantic judgments that depend on study context. Programs use Code Logic to validate identifiers, perform numerical computations, verify sources, execute formalized methodological rules, and deterministically validate the corresponding outputs. Thus, Task Contracts define task constraints and allocate responsibilities, while Code Logic executes the constraints that can be formalized.

To maintain correspondence between outputs and original evidence, MetaRigor uses Source Anchors to bind outputs to retrievable source locations. Each Source Anchor stores the source identifier, source location, retrieved text, and hash values. For task $j$, let $\mathcal{D}_j$ denote the set of permitted source identifiers, and let $c$ denote a citation submitted by the model. Parse resolves the citation into a source identifier $d$ and location $\ell$, while Read retrieves the corresponding source text $q$:

\begin{equation}
B_{\mathcal{D}_j}(c)=
\begin{cases}
(d,\ell,q), &
\begin{aligned}
& \text{if } \operatorname{Parse}(c)=(d,\ell), \\
& d\in\mathcal{D}_j, \\
& q=\operatorname{Read}(d,\ell)\neq\bot,
\end{aligned} \\
\bot, & \text{otherwise}.
\end{cases}
\end{equation}

\begin{figure}[ht]
    \centering
    \includegraphics[width=\linewidth]{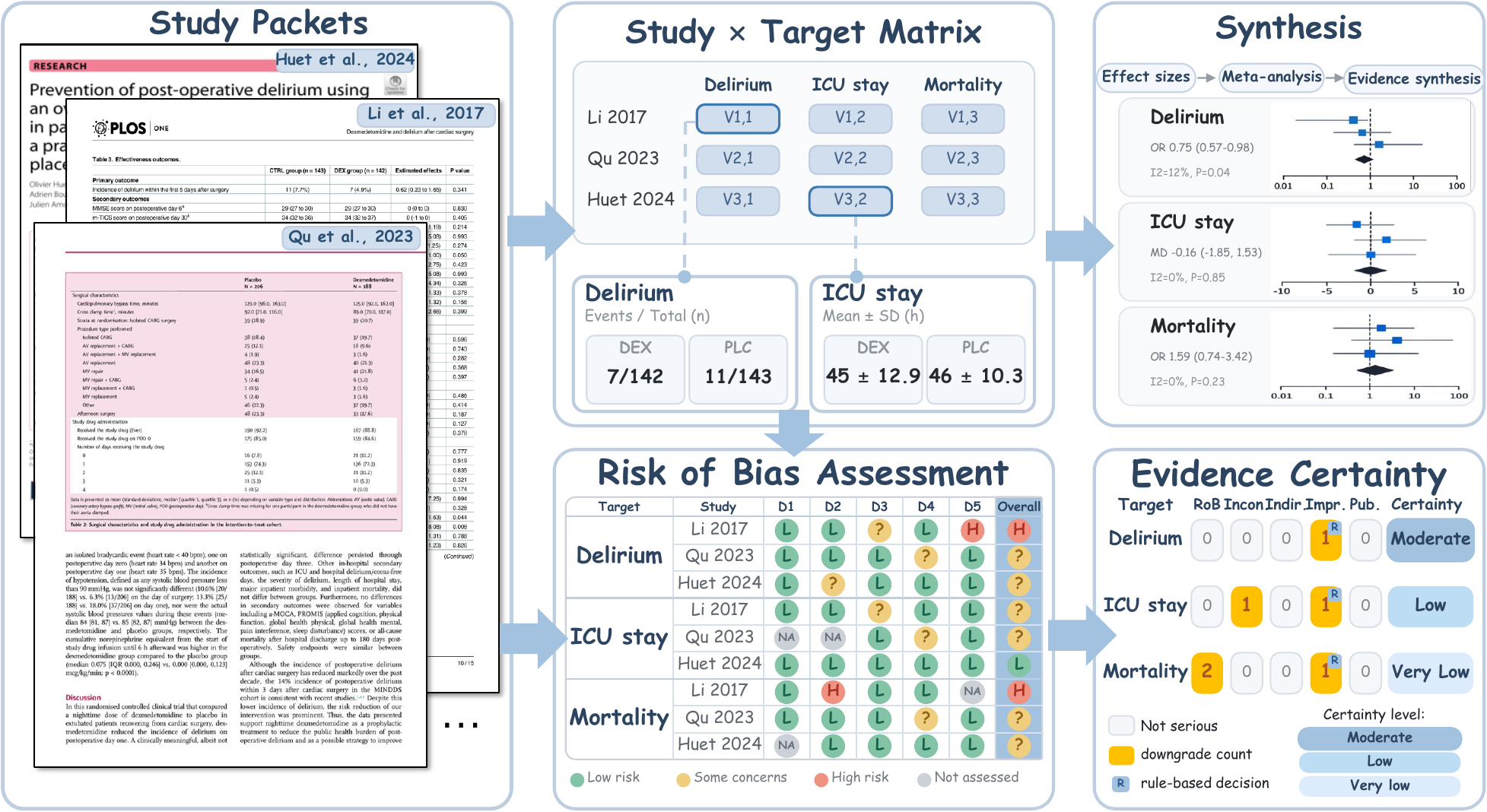}
    \caption{Organization and provenance tracking of evidence synthesis artifacts in MetaRigor. Source materials are consolidated into Study $\times$ Target results and propagated to statistical synthesis, risk of bias assessment, and certainty of evidence assessment. The figure illustrates evidence consolidation at the study level, result alignment, and downstream synthesis.}
    \label{fig:downstream}
\end{figure}

Here, $\bot$ denotes a source binding failure. Parsing failures, violations of the permitted source boundary, and retrieval failures are all recorded as issues. The Verifier further checks citation formatting and deterministic constraints while preserving the associated source information. The system distinguishes technical source closure, measured by Claim Closure, from evidence level recall, measured by Reference Recall. The former verifies whether an output is linked to a valid and retrievable source location, whereas the latter evaluates whether retrieved evidence overlaps with manually annotated supporting evidence spans (Figure~\ref{fig:downstream}).

\subsection{Five Stage Evidence Synthesis Workflow}

MetaRigor organizes its five stages according to the practical execution order of evidence synthesis. All stages share the research representation and task execution mechanisms described above. In evidence acquisition (EA), the system constructs the materials for analysis through literature screening and study consolidation. To reduce dependence on any single model's judgments, the system employs two models with comparable task capabilities but distinct judgment tendencies to independently assess each record. Both models use the same Protocol, inclusion and exclusion criteria, and output requirements, without sharing their intermediate judgments. Maintaining independent judgments allows the system to identify potential disagreements between models regarding study eligibility and resolve inconsistencies through predefined rules. At the title and abstract stage, a record proceeds to full-text assessment only when both models successfully complete their assessments and explicitly determine that the record is eligible. At the full-text stage, the two models again make independent judgments. When their judgments agree, the screening outcome is determined directly. Disagreements are referred to an independent adjudication model. After screening, the unit of organization shifts from reports to studies. Primary reports, supplementary materials, registry records, and subsequent materials are linked to the corresponding Study Packet. Reports whose study identities cannot be established are retained as unresolved.

In data extraction (DE), MetaRigor separates the extraction of observations actually reported in the sources from the selection of results guided by the Protocol. An Observation records the comparison, outcome, time point, population, analytical conditions, group roles, numerical values, and source information reported in the original material. Result selection determines whether an observation is compatible with the specified Target and records the rationale for its acceptance or rejection. For binary outcomes, the delivered result contains event counts and denominators for each group. For continuous outcomes, it contains the mean, standard deviation, and sample size for each group. Every adopted numerical field must be linked to corresponding evidence in the original text. When information is insufficient, sources conflict, or a unique candidate result cannot be identified, the system retains partial results and ambiguity rather than combining information across time points or analysis populations. The validated data are subsequently passed to statistical tools in accordance with the statistical analysis plan specified in the Protocol. These tools calculate effect sizes, estimate confidence intervals, pool effects across studies, and analyze heterogeneity.

In risk of bias assessment (RoB), the system uses the August 22, 2019 version of RoB 2, the Cochrane tool for assessing risk of bias in randomized trials, to assess the effect of assignment to intervention in individually randomized parallel group trials \citep{sterne2019rob}. RoB 2 comprises five domains. These address bias arising from the randomization process (D1), bias due to deviations from intended interventions (D2), bias due to missing outcome data (D3), bias in measurement of the outcome (D4), and bias in selection of the reported result (D5). D1 is assessed at the study level, D2 is linked to the intervention effect under assessment, and D3 through D5 are linked to the specific result. This mapping preserves the Study $\times$ Target identities established above. The model answers signaling questions and selects supporting evidence, while programs execute the applicable decision branches and domain rules.

Overall risk is aggregated according to programmatic rules. If any domain is rated High risk, the overall rating is High risk. If all domains are rated Low risk, the overall rating is Low risk. All other cases are rated Some concerns. The current implementation does not yet cover context dependent judgments in which multiple domains rated Some concerns may jointly warrant an overall High risk rating. Responses to signaling questions, domain judgments, and the overall judgment are all delivered with their supporting sources.

In certainty of evidence assessment (EC), the system applies GRADE to evaluate five domains for a body of evidence defined by a fixed comparison, outcome, and other conditions prespecified in the Protocol. This stage assesses the body of evidence synthesized across multiple studies for the same target, rather than independently evaluating individual Study $\times$ Target pairs. The model proposes downgrading judgments for risk of bias, inconsistency, indirectness, and publication bias. Imprecision is determined programmatically using predefined rules. The current imprecision branch is designed for binary outcomes expressed as RR or OR and applies rules based on the confidence interval, prespecified thresholds, and the optimal information size (OIS) \citep{guyatt2025core}.

Overall certainty is generated through semantic synthesis under explicit interval constraints. After all five domain judgments are completed, the program maps each domain label to a corresponding downgrade interval $[\lambda_i^{-},\lambda_i^{+}]$ and defines:

\begin{equation}
L=\max_{1\leq i\leq 5}\lambda_i^{-}, \qquad U=\min\left(3,\sum_{i=1}^{5}\lambda_i^{+}\right), \qquad \Delta\in[L,U]\cap\mathbb{Z}.
\end{equation}

The lower bound $L$ preserves the largest minimum downgrading requirement among the five domains, while the upper bound $U$ aggregates the maximum permitted downgrading across domains and caps the total at three levels. The EC synthesis model uses domain level rationales to identify potential duplicate downgrading and proposes an integer $\Delta$ within the admissible interval. The program then verifies that the proposed value satisfies the constraints. Starting from High certainty, $\Delta=0,1,2,3$ correspond to High, Moderate, Low, and Very low certainty, respectively. The interval constraint restricts the range of the final downgrade count but does not replace the model's semantic judgment about duplicate downgrading. Both domain judgments and the overall certainty assessment retain their source relationships, allowing them to be traced back to study materials, RoB outputs, and statistical artifacts, thereby supporting end-to-end evidence traceability.

In manuscript generation (MS), the system does not reassess scientific facts established in upstream stages. Instead, it organizes existing research artifacts within explicit fact scopes and generates the manuscript in accordance with the PRISMA 2020 reporting guidelines \citep{page2021prisma}, while maintaining consistency with upstream evidence. The model first develops a plan for the manuscript, then drafts the Introduction, Methods, Results, and Discussion as appropriate based on the Protocol, execution records, and study findings. Finally, it generates the title and abstract.

\section{Experiment}

\subsection{Experimental Setup}

We conducted controlled evaluations across five stages to assess MetaRigor in terms of study material recovery, target result delivery, methodological judgments, manuscript factual fidelity, and source verifiability. Reference annotations were independently produced by two researchers, with disagreements resolved through discussion. Each stage used fixed upstream inputs to distinguish execution differences at the current stage from errors propagated from preceding stages. Input materials and inter-annotator agreement are provided in Appendix \ref{app:A}.

Under identical research questions, stage inputs, and material conditions, we compared MetaRigor against four baseline systems. Codex CLI and OpenHands CLI served as general-purpose agent baselines, while Manalyzer and meta-pipe served as domain-specific baselines. Manalyzer participated only in EA and DE, whereas the other three baselines participated in all five stages. Model, tool, and execution configurations are detailed in Appendix \ref{app:E}.

We additionally conducted a simulated end-to-end evaluation on
seven cases, comparing MetaRigor with Codex CLI, OpenHands CLI,
and meta-pipe. Starting from the data extraction stage, each
system proceeded through the subsequent assessment, synthesis,
and manuscript-generation stages. We evaluated factual correctness and coverage in the final
manuscript package against a shared reference set. The package
comprised the DE, RoB, EC, and MS artifacts. Evaluation criteria and execution
configurations are detailed in Appendix~\ref{app:G} to support
transparent and reproducible evaluation.

For the ablation experiments, the original EA configuration uses DeepSeek-V4-Flash and GPT-5.6 Luna to execute the two independent screening paths, respectively. To examine the effect of model configuration across the two screening paths, we further evaluate two configurations, Dual-Luna and Dual-DeepSeek, which use GPT-5.6 Luna and DeepSeek-V4-Flash, respectively, for both screening paths. The remaining stages evaluated two configurations, w/o Task Contract and w/o Code Logic. The former removed the explicit Task Contract, while the latter transferred selected formalizable rules from programmatic execution to model-based judgment. Metric definitions, denominators, and scoring procedures are provided in Appendix \ref{app:D}.

\subsection{Comparative Experiments}
\label{sec:comparative}



\begin{table}[htbp]
\centering
\caption[Comparative performance in evidence acquisition, data extraction, risk of bias assessment, and certainty of evidence assessment under identical evaluation settings. Best and second-best results are indicated in bold and underlined, respectively.]{%
\protect\hl{Comparative performance in evidence acquisition, data extraction, risk of bias assessment, and certainty of evidence assessment under identical evaluation settings. Best and second-best results are indicated in bold and underlined, respectively.}%
}
\label{tab:comparative_performance}

\begingroup
\setlength{\tabcolsep}{3.3pt}
\renewcommand{\arraystretch}{1.10}

\begin{adjustbox}{max width=\textwidth,center}
\begin{tabular}{@{}l@{\hspace{3pt}}cccc ccc ccc@{}}
\toprule
& \multicolumn{4}{c}{\textbf{Data Extraction (DE)}}
& \multicolumn{3}{c}{\textbf{Risk of Bias Assessment (RoB)}}
& \multicolumn{3}{c}{\textbf{Evidence Certainty (EC)}} \\
\cmidrule(lr){2-5}
\cmidrule(lr){6-8}
\cmidrule(lr){9-11}

\textbf{Method}
& \textbf{F\textsubscript{1}}$\uparrow$
& \textbf{Recall}$\uparrow$
& \textbf{Field Acc.}$\uparrow$
& \textbf{Precision}$\uparrow$
& \textbf{Domain}$\uparrow$
& \textbf{Judgment}$\uparrow$
& \textbf{High Risk}$\uparrow$
& \textbf{Domain}$\uparrow$
& \textbf{Downgrade}$\uparrow$
& \textbf{MAE}$\downarrow$ \\
\midrule

Codex
& \underline{84.62\%}
& \underline{84.62\%}
& \underline{85.99\%}
& 84.62\%
& \underline{68.13\%}
& \underline{59.38\%}
& 30.77\%
& \underline{62.20\%}
& \underline{45.00\%}
& \underline{0.590} \\

OpenHands
& 74.87\%
& 71.79\%
& 77.66\%
& 78.21\%
& 60.00\%
& 56.25\%
& \underline{46.15\%}
& 60.60\%
& 35.00\%
& 0.727 \\

\hl{Manalyzer}
& 48.65\%
& 32.31\%
& 35.75\%
& \textbf{98.44\%}
& N/A
& N/A
& N/A
& N/A
& N/A
& N/A \\

\hl{meta-pipe}
& 53.02\%
& 40.51\%
& 44.93\%
& 76.70\%
& 64.38\%
& 50.00\%
& 7.69\%
& 57.80\%
& 31.00\%
& 0.990 \\

\textbf{MetaRigor}
& \textbf{86.29\%}
& \textbf{87.18\%}
& \textbf{88.89\%}
& \underline{85.43\%}
& \textbf{71.25\%}
& \textbf{62.50\%}
& \textbf{69.23\%}
& \textbf{80.20\%}
& \textbf{55.00\%}
& \textbf{0.413} \\

\bottomrule
\end{tabular}
\end{adjustbox}

\par\vspace{3pt}

\begin{adjustbox}{max width=\textwidth,center}
\begin{tabular}{@{}l@{\hspace{3pt}}cccc cccc cc@{}}
& \multicolumn{4}{c}{\textbf{Title/Abstract Stage (EA)}}
& \multicolumn{4}{c}{\textbf{Full-Text Stage (EA)}}
& \multicolumn{2}{c}{\textbf{Overall (EA)}} \\
\cmidrule(lr){2-5}
\cmidrule(lr){6-9}
\cmidrule(lr){10-11}

\textbf{Method}
& \textbf{Precision}$\uparrow$
& \textbf{F\textsubscript{1}}$\uparrow$
& \textbf{F\textsubscript{2}}$\uparrow$
& \textbf{NNR}$\downarrow$
& \textbf{Precision}$\uparrow$
& \textbf{F\textsubscript{1}}$\uparrow$
& \textbf{F\textsubscript{2}}$\uparrow$
& \textbf{NNR}$\downarrow$
& \textbf{Family Rec.}$\uparrow$
& \textbf{Comp. Rec.}$\uparrow$ \\
\midrule

Codex
& 15.34\%
& 26.60\%
& 47.53\%
& 6.52
& 18.90\%
& 31.58\%
& 52.86\%
& 5.29
& 32.00\%
& 79.31\% \\

OpenHands
& 20.78\%
& 31.37\%
& 45.20\%
& 4.81
& 51.02\%
& 67.57\%
& \underline{83.89\%}
& 1.96
& \underline{60.00\%}
& 31.03\% \\

\hl{Codex}\textsuperscript{\(\dagger\)}
& 14.62\%
& 25.51\%
& 46.13\%
& 6.84
& 53.49\%
& 67.65\%
& 80.42\%
& 1.87
& 28.00\%
& 44.83\% \\

\hl{OpenHands}\textsuperscript{\(\dagger\)}
& 9.64\%
& 17.52\%
& 34.38\%
& 10.38
& \underline{57.50\%}
& \underline{70.77\%}
& 82.14\%
& \underline{1.74}
& 32.00\%
& 17.24\% \\

\hl{Manalyzer}
& \underline{44.23\%}
& \underline{59.74\%}
& \underline{75.66\%}
& \underline{2.26}
& 45.10\%
& 60.53\%
& 76.16\%
& 2.22
& 36.00\%
& 27.59\% \\

\hl{meta-pipe}
& 18.80\%
& 31.65\%
& 53.65\%
& 5.32
& 20.66\%
& 34.25\%
& 56.56\%
& 4.84
& 48.00\%
& \underline{82.76\%} \\

\textbf{MetaRigor}
& \textbf{52.17\%}
& \textbf{67.61\%}
& \textbf{82.19\%}
& \textbf{1.92}
& \textbf{63.16\%}
& \textbf{76.19\%}
& \textbf{86.96\%}
& \textbf{1.58}
& \textbf{88.00\%}
& \textbf{93.10\%} \\

\bottomrule
\end{tabular}
\end{adjustbox}

\par\vspace{3pt}
\begin{minipage}{\linewidth}
\footnotesize
\textsuperscript{\(\dagger\)}
These variants use GPT-5.6 Luna throughout the EA workflow.
\end{minipage}

\endgroup
\end{table}

Table~\ref{tab:comparative_performance} summarizes the comparative
results across four stages. MetaRigor achieved the best performance
on most primary metrics. In DE, its $F_1$, recall, and field accuracy
reached 86.29\%, 87.18\%, and 88.89\%, respectively.
Although Manalyzer achieved 98.44\% precision, its recall was only
32.31\%. In RoB, MetaRigor achieved 69.23\% high-risk recall,
compared with 30.77\% for Codex CLI, 46.15\% for OpenHands CLI,
and 7.69\% for meta-pipe. In EC, its domain agreement reached
80.20\%, with a downgrade MAE of 0.413.

In EA, MetaRigor achieved $F_1$ scores of 67.61\% and 76.19\%
at the title and abstract and full-text stages, respectively,
with corresponding $F_2$ scores of 82.19\% and 86.96\%.
It also achieved the lowest screening burden at the two screening stages, with NNR values
of 1.92 and 1.58, respectively, recovering 88.00\% of study
families and 93.10\% of companion reports.
Its performance also exceeded that of the CLI baselines
using GPT-5.6 Luna.

\begin{figure}[ht]
    \centering
    \includegraphics[width=\linewidth]{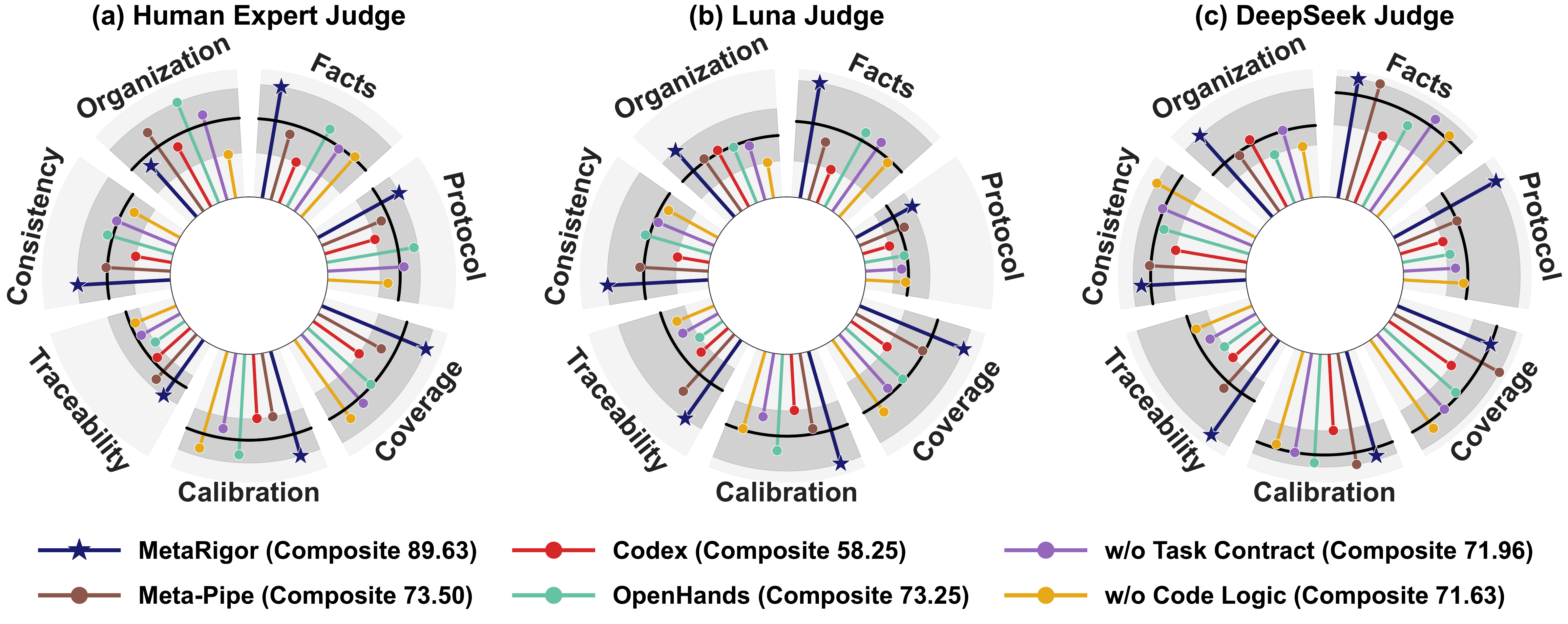}
    \caption[Evaluation of manuscript generation quality under comparative and ablation settings. Human experts, Luna, and DeepSeek assess generated manuscripts across seven criteria, including factual accuracy, Protocol adherence, material coverage, and traceability.]{%
\protect\hl{Evaluation of manuscript generation quality under comparative and ablation settings. Human experts, Luna, and DeepSeek assess generated manuscripts across seven criteria, including factual accuracy, Protocol adherence, material coverage, and traceability.}%
}
    \label{fig:judge}
\end{figure}

In the human evaluation of MS, MetaRigor scored 86.00,
outperforming OpenHands CLI at 77.50, meta-pipe at 68.13,
and Codex CLI at 57.25, as shown in Figure \ref{fig:judge}.
Luna and DeepSeek evaluations likewise ranked MetaRigor first,
assigning scores of 88.25 and 94.63, respectively.
Its advantages were most evident in factual fidelity,
Protocol adherence, material coverage, and source traceability. An additional model-controlled comparison using GPT-6.1 Terra
is presented in Appendix~\ref{app:terra}.

\subsection{Ablation Experiments}
\label{sec:ablation}
Ablation results in Table~\ref{tab:ablation_study} demonstrate
the contributions of both Task Contract and Code Logic
to system performance.
Removing Task Contract reduced DE $F_1$ from 86.29\% to 81.61\%,
RoB domain agreement from 71.25\% to 48.75\%, and the MS composite
score from 89.63 to 71.96.
Removing Code Logic reduced RoB high-risk recall from 69.23\%
to 15.38\%, increased EC downgrade MAE from 0.413 to 0.600,
and reduced the MS composite score to 71.63
(Figure~\ref{fig:judge}).

In EA, Dual Luna and Dual DeepSeek achieved title and abstract
$F_1$ scores of 62.34\% and 22.94\%, and full-text $F_1$ scores
of 70.97\% and 58.67\%, respectively.
The full configuration achieved 67.61\% and 76.19\%, respectively,
along with the highest $F_2$ scores and study recovery rates.
These results indicate that screening performance is sensitive
to model configuration, with the full configuration achieving
the best overall performance in our evaluation.

\vspace{2mm}


\begin{table}[ht]
\centering

\caption[Ablation analysis of key design components in MetaRigor. The table evaluates the contribution of task contracts, code logic, and stage-specific components across evidence acquisition (EA), data extraction (DE), risk-of-bias assessment (RoB), and evidence certainty assessment (EC).]{%
\protect\hl{Ablation analysis of key design components in MetaRigor. The table evaluates the contribution of task contracts, code logic, and stage-specific components across evidence acquisition (EA), data extraction (DE), risk-of-bias assessment (RoB), and evidence certainty assessment (EC).}%
}
\label{tab:ablation_study}

\setlength{\tabcolsep}{3.3pt}
\renewcommand{\arraystretch}{1.10}


\begin{adjustbox}{max width=\textwidth,center}
\begin{tabular}{@{}l@{\hspace{3pt}}cccc ccc ccc@{}}

\toprule

&
\multicolumn{4}{c}{\textbf{Data Extraction (DE)}}
&
\multicolumn{3}{c}{\textbf{Risk of Bias Assessment (RoB)}}
&
\multicolumn{3}{c}{\textbf{Evidence Certainty (EC)}}
\\

\cmidrule(lr){2-5}
\cmidrule(lr){6-8}
\cmidrule(lr){9-11}

\textbf{Method}
& \textbf{Recall$\uparrow$}
& \textbf{F$_1$$\uparrow$}
& \textbf{Field Acc.$\uparrow$}
& \textbf{Precision$\uparrow$}
& \textbf{Domain$\uparrow$}
& \textbf{Judgment$\uparrow$}
& \textbf{High Risk$\uparrow$}
& \textbf{Domain$\uparrow$}
& \textbf{Downgrade$\uparrow$}
& \textbf{MAE$\downarrow$}
\\

\midrule

w/o Task Contract
& \secondbest{83.08\%}
& \secondbest{81.61\%}
& \secondbest{86.84\%}
& \secondbest{80.20\%}
& 48.75\%
& \secondbest{43.75\%}
& \secondbest{38.46\%}
& \secondbest{70.60\%}
& 45.00\%
& \secondbest{0.450}
\\

w/o Code Logic
& N/A
& N/A
& N/A
& N/A
& \secondbest{61.88\%}
& 18.75\%
& 15.38\%
& 68.80\%
& \secondbest{46.00\%}
& 0.600
\\

\textbf{MetaRigor}
& \best{87.18\%}
& \best{86.29\%}
& \best{88.89\%}
& \best{85.43\%}
& \best{71.25\%}
& \best{62.50\%}
& \best{69.23\%}
& \best{80.20\%}
& \best{55.00\%}
& \best{0.413}
\\

\bottomrule

\end{tabular}
\end{adjustbox}

\par\vspace{3pt}


\begin{adjustbox}{max width=\textwidth,center}
\begin{tabular}{@{}l@{\hspace{3pt}}cccc cccc cc@{}}

&
\multicolumn{4}{c}{\textbf{Title/Abstract Stage (EA)}}
&
\multicolumn{4}{c}{\textbf{Full-Text Stage (EA)}}
&
\multicolumn{2}{c}{\textbf{Overall (EA)}}
\\

\cmidrule(lr){2-5}
\cmidrule(lr){6-9}
\cmidrule(lr){10-11}

\textbf{Method}
& \textbf{Precision$\uparrow$}
& \textbf{F$_1$$\uparrow$}
& \textbf{F$_2$$\uparrow$}
& \textbf{NNR$\downarrow$}
& \textbf{Precision$\uparrow$}
& \textbf{F$_1$$\uparrow$}
& \textbf{F$_2$$\uparrow$}
& \textbf{NNR$\downarrow$}
& \textbf{Family Rec.$\uparrow$}
& \textbf{Comp. Rec.$\uparrow$}
\\

\midrule

\hl{Dual Luna}
& \secondbest{46.15\%}
& \secondbest{62.34\%}
& \secondbest{78.95\%}
& \secondbest{2.17}
& \secondbest{59.46\%}
& \secondbest{70.97\%}
& \secondbest{80.29\%}
& \secondbest{1.68}
& \secondbest{84.00\%}
& 79.31\%
\\

Dual DeepSeek
& 12.95\%
& 22.94\%
& 42.66\%
& 7.72
& 44.00\%
& 58.67\%
& 73.33\%
& 2.27
& \secondbest{84.00\%}
& \secondbest{86.21\%}
\\

\textbf{MetaRigor}
& \best{52.17\%}
& \best{67.61\%}
& \best{82.19\%}
& \best{1.92}
& \best{63.16\%}
& \best{76.19\%}
& \best{86.96\%}
& \best{1.58}
& \best{88.00\%}
& \best{93.10\%}
\\

\bottomrule

\end{tabular}
\end{adjustbox}

\end{table}

\subsection{ End-to-End Comparative Evaluation}

\begin{table}[htbp]
\centering
\caption[Micro-averaged factual performance of the final manuscript package.]{%
\protect\hl{Micro-averaged factual performance of the final manuscript package.}%
}
\label{tab:system_results}

\begingroup
\small
\setlength{\tabcolsep}{5pt}
\renewcommand{\arraystretch}{1.15}

\begin{tabularx}{\linewidth}{
  @{}l*{6}{>{\centering\arraybackslash}X}@{}
}
\toprule
\textbf{Method}
& \textbf{Correct}
& \textbf{Predicted}
& \textbf{Reference}
& \textbf{F\textsubscript{1}}$\uparrow$
& \textbf{Precision}$\uparrow$
& \textbf{Recall}$\uparrow$ \\
\midrule

meta-pipe
& 814
& 1235
& 2133
& 48.34\%
& 65.91\%
& 38.16\% \\

Codex
& 1310
& 1764
& 2133
& 67.23\%
& 74.26\%
& 61.42\% \\

OpenHands
& 970
& 1610
& 2133
& 51.83\%
& 60.25\%
& 45.48\% \\

\textbf{MetaRigor}
& 1539
& 1864
& 2133
& \textbf{77.01\%}
& \textbf{82.56\%}
& \textbf{72.15\%} \\

\bottomrule
\end{tabularx}

\endgroup
\end{table}

Across seven cases evaluated using a shared reference set,
MetaRigor correctly reported 1,539 of 2,133 reference facts
across the delivered DE, RoB, EC, and MS artifacts,
with a micro-averaged factual precision of 82.56\%,
recall of 72.15\%, and $F_1$ score of 77.01\%,
all higher than those of other systems
(Table~\ref{tab:system_results}).
Compared with Codex CLI, OpenHands CLI, and meta-pipe,
MetaRigor achieved $F_1$ improvements of 9.78, 25.18,
and 28.67 percentage points, respectively.

In two illustrative MetaRigor runs, corresponding to the Lin and Turalde cases, the generated artifacts consistently preserved Study $\times$ Target identities, source bindings, and the status of unresolved items and missing information. We further traced the outputs for Lin et al. (2025), a systematic review of endovascular thrombectomy for large ischemic strokes. SELECT2 was a randomized controlled trial comparing endovascular thrombectomy plus medical therapy with medical therapy alone. For the prespecified target of all-cause mortality at 90 days, OpenHands CLI incorrectly assigned the in-hospital mortality data (42/178 vs. 44/174) to this target, whereas MetaRigor selected the correct 90-day mortality data (68/177 vs. 71/171). Furthermore, in the RESCUE-Japan LIMIT trial included in the same review, meta-pipe reported 58 events in the intervention group for the 90-day outcome of mRS scores of 5 or 6, compared with the reference value of 36. This error was subsequently propagated to statistical calculations and manuscript generation. These cases illustrate how mismatches in result identity and numerical errors can propagate to downstream stages. Detailed artifact comparisons and analyses are provided in Appendix \ref{app:G}.

\section{Conclusion}

In this work, we presented MetaRigor, a framework for verifiable end-to-end evidence synthesis built around explicit Scientific Decision Boundaries. MetaRigor maintains Study $\times$ Target identities across stages, uses Task Contracts to separate model interpretation from programmatic rule execution, and binds outputs to source evidence. Controlled evaluations across five stages show improvements over four baselines in evidence acquisition, complete result delivery, methodological assessment, and manuscript fidelity. A seven-case downstream end-to-end evaluation yielded an aggregate $F_1$ score of 77.01\%, exceeding the strongest baseline by 9.78 percentage points. Ablations identify distinct, stage-dependent contributions from task constraints and programmatic rules. Tracing archived outputs reveals why cross-stage consistency matters: correctly extracted values can enter final reports under the wrong outcome or time point, compromising downstream synthesis. While broader validation remains necessary, our results motivate scientific agents that explicitly preserve the relationships among evidence, decisions, and results throughout execution.

\subsection*{AI use statement}

We used generative AI tools solely for manuscript refinement, including language translation, grammar correction, and improving clarity and readability. AI tools were not used for research ideation, hypothesis generation, experimental design, data analysis, result generation, dataset creation, mathematical derivations, proofs, or scientific claims. All AI-assisted suggestions were reviewed and verified by the authors, who take full responsibility for the final content.

\subsection*{Ethics statement}

MetaRigor is designed to support, rather than replace, human judgment in evidence synthesis and is not intended to provide clinical recommendations. Automated evidence acquisition, data extraction, methodological assessment, and manuscript generation may introduce errors or unsupported interpretations; therefore, system outputs should be independently verified by researchers. The experiments use existing scholarly materials and research artifacts and do not involve the recruitment of human participants or the collection of new personally identifiable information. Responsibility for study inclusion, methodological judgments, evidence interpretation, and downstream scientific or clinical claims remains with human researchers.

\subsection*{Reproducibility Statement}

Source code and evaluation artifacts are available at
\url{https://anonymous.4open.science/r/Meta-Rigor-8C62}.

\bibliography{iclr2027_conference}
\bibliographystyle{iclr2027_conference}

\appendix

\counterwithout{equation}{section}
\renewcommand{\theequation}{\arabic{equation}}

\section{Reference Sets and Stage Inputs}
\label{app:A}

\subsection{Construction of Reference Answers}

Reference answers were independently annotated by two researchers, with disagreements resolved through discussion. EA counts followed the report- and study-grouping conventions of the original reviews and are not listed separately in the agreement table. DE was evaluated at the Study $\times$ Target unit and numerical-field levels; RoB annotations covered signaling questions, domain judgments, and overall judgments; and EC annotations covered judgments across five domains and overall judgments. Table \ref{tab:app_agreement} presents the annotation counts and pre-adjudication agreement at each evaluation level under the same evaluation protocol.

\begin{table}[ht]
  \centering
  \caption{Pre-adjudication agreement of reference annotations.}
  \label{tab:app_agreement}

  \begingroup
  \small
  \color{black}
  \setlength{\tabcolsep}{2pt}
  \setlength{\aboverulesep}{1.5pt}
  \setlength{\belowrulesep}{1.5pt}
  \renewcommand{\arraystretch}{1.15}

  \begin{tabular*}{\linewidth}
    {@{\extracolsep{\fill}}lccccc@{}}
    \toprule
    \textbf{Stage}
    & \textbf{Evaluation unit}
    & \textbf{\shortstack{Annotation units}}
    & \textbf{\shortstack{Initial\\matches}}
    & \textbf{\shortstack{Initial\\agreement}}
    & \textbf{Discussed} \\
    \midrule

    \multirow{2}{*}{DE}
    & Study $\times$ Target
    & 405 & 393 & 97.0\% & 12 \\
    & Numerical fields
    & 828 & 816 & 98.6\% & 12 \\

    \midrule
    RoB
    & \shortstack{Signaling questions /\\domains / overall}
    & 896 & 858 & 95.8\% & 38 \\

    \midrule
    EC
    & Five domains / overall
    & 600 & 567 & 94.5\% & 33 \\

    \bottomrule
  \end{tabular*}
  \endgroup
\end{table}


\subsection{Stage Inputs and Evaluation References}

The stage$-$controlled evaluations in Sections~\ref{sec:comparative} and~\ref{sec:ablation} of the main text use fixed upstream inputs to distinguish differences within the current stage from upstream error propagation. Table \ref{tab:app_inputs} lists the task-visible materials and evaluation references for each stage under the same evaluation protocol. Inputs do not include the final answers to be predicted at the current stage; upstream results required by downstream tasks are provided only as permitted facts. Reviews reused across stages are not counted again as new, independent review samples.

\begin{table}[ht]
  \caption{Inputs and references for stage-controlled evaluation.}
  \label{tab:app_inputs}
  \centering
  \begingroup
  \small
  \color{black}
  \setlength{\tabcolsep}{2pt}
  \setlength{\aboverulesep}{1.5pt}
  \setlength{\belowrulesep}{1.5pt}
  \renewcommand{\arraystretch}{1.15}

  \begin{tabular*}{\linewidth}
    {@{\extracolsep{\fill}}lp{0.38\linewidth}p{0.38\linewidth}}
    \toprule
    \textbf{Stage}
    & \multicolumn{1}{c}{\textbf{Task-visible materials}}
    & \multicolumn{1}{c}{\textbf{Evaluation reference}} \\
    \midrule

    EA
    & Protocol, bibliographic records or full texts, and report relationships
    & Reference study set and known report families \\

    DE
    & Protocol, frozen source representations, and Study $\times$ Target list
    & Study $\times$ Target numerical results and statuses \\

    RoB
    & Study materials, result identity, and signaling questions
    & Signaling-question, domain, and overall labels \\

    EC
    & Comparisons, outcomes, RoB results, and statistical artifacts
    & Five-domain judgments and overall downgrade counts \\

    MS
    & Protocol, execution records, and research findings
    & Seven-dimensional scoring rubric and permitted facts \\

    \bottomrule
  \end{tabular*}
  \endgroup
\end{table}


\subsection{Task Scale and Scope of Supplementary Analyses}

Table~\ref{tab:app_task_scale} summarizes the scale of the five-stage controlled evaluation across all stage-specific tasks. \hl{Within each stage, each case corresponds to one distinct systematic review. The same review may be evaluated across multiple stages; therefore, stage-specific case counts are reported separately and should not be summed to estimate the total number of distinct reviews across stages.}

\begin{table}[ht]
  \caption{Task scale of the five-stage controlled evaluation.}
  \label{tab:app_task_scale}
  \centering
  \begingroup
  \small
  \color{black}
  \setlength{\tabcolsep}{2pt}
  \setlength{\aboverulesep}{1.5pt}
  \setlength{\belowrulesep}{1.5pt}
  \renewcommand{\arraystretch}{1.15}

  \begin{tabular*}{\linewidth}
    {@{\extracolsep{\fill}}lcc@{}}
    \toprule
    \textbf{Stage}
    & \textbf{Cases}
    & \textbf{Evaluation scale} \\
    \midrule

    EA & 5
    & 822 title--abstract records; 300 full texts; 25 reference studies and 29 companion reports \\

    DE & 10
    & 405 Study $\times$ Target units; 195 reported results; 828 numerical fields \\

    RoB & 7
    & 704 signaling questions; 32 study--result units; 160 domain judgments \\

    EC & 100
    & 100 fixed analyses; 500 domain judgments; 100 overall downgrade judgments \\

    MS & 10
    & 10 manuscripts per condition; 7 scoring criteria per manuscript \\

    \bottomrule
  \end{tabular*}
  \endgroup
\end{table}

Appendix \ref{app:F} provides tool-call traces from the EC ablation run, and Appendix \ref{app:G} documents continuous-execution states, archived artifacts from the four systems, and downstream artifact displays. Case identifiers and audit details are provided only in the corresponding appendices. These supplementary analyses are neither pooled with the stage-controlled evaluations to calculate accuracy nor counted as additional independent review samples.

\subsection{Systematic Review Sources and Dataset Composition}
\label{app:review_sources}

The experiments covered 102 cases, each corresponding to a distinct
systematic review. Each review included 2 to 29 primary studies,
yielding a total of 717 review level study inclusions across all
reviews (Table~\ref{tab:review_sources}).

\begingroup
\small
\setlength{\tabcolsep}{4pt}
\renewcommand{\arraystretch}{1.15}
\setlength{\LTleft}{0pt}
\setlength{\LTright}{0pt}

\begin{longtable}{
  @{}
  >{\raggedright\arraybackslash}p{0.07\linewidth}
  >{\raggedright\arraybackslash}p{\dimexpr0.34\linewidth-8pt\relax}
  >{\centering\arraybackslash}p{\dimexpr0.18\linewidth-8pt\relax}
  >{\centering\arraybackslash}p{\dimexpr0.26\linewidth-8pt\relax}
  >{\centering\arraybackslash}p{\dimexpr0.15\linewidth-8pt\relax}
  @{}
}

\caption[Sources, evaluation stages, and numbers of included studies for the systematic reviews used in the experiments.]{%
\protect\hl{Sources, evaluation stages, and numbers of included studies for the systematic reviews used in the experiments.}%
}
\label{tab:review_sources}\\

\toprule
\textbf{Case}
& \textbf{Author and Year}
& \textbf{Review PMCID}
& \textbf{Stages}
& \textbf{Included Studies (n)} \\
\midrule
\endfirsthead

\multicolumn{5}{l}{
\footnotesize Table~\thetable\ (continued)
}\\
\toprule
\textbf{Case}
& \textbf{Author and Year}
& \textbf{Review PMCID}
& \textbf{Stages}
& \textbf{Included Studies (n)} \\
\midrule
\endhead

\midrule
\multicolumn{5}{r}{\footnotesize Continued on next page}\\
\endfoot

\bottomrule
\endlastfoot

R001 & Meng et al., 2025
& PMC11783797 & EA, DE, EC, MS & 7 \\

R002 & Wang et al., 2025
& PMC12440473 & DE, RoB, EC, MS & 9 \\

R003 & Turalde-Mapili et al., 2023
& PMC10110918 & EA, DE, RoB, EC, MS & 2 \\

R004 & Zhong et al., 2022
& PMC9334708 & EA, DE, EC, MS & 5 \\

R005 & Song et al., 2025
& PMC12533610 & DE, RoB, EC, MS & 4 \\

R006 & Lin et al., 2025
& PMC12037071 & EA, DE, RoB, EC, MS & 6 \\

R007 & Malhotra et al., 2025
& PMC12757263 & DE, MS & 2 \\

R008 & Balbaa et al., 2025
& PMC11869692 & EA, DE, RoB, MS & 5 \\

R009 & Sun et al., 2024
& PMC11221214 & DE, RoB, EC, MS & 5 \\

R010 & Qazi et al., 2026
& PMC13257171 & DE, RoB, EC, MS & 5 \\

R011 & Toti et al., 2026
& PMC13290216 & EC & 10 \\

R012 & Yang et al., 2026
& PMC13337450 & EC & 24 \\

R013 & Khalid et al., 2026
& PMC13348439 & EC & 7 \\

R014 & Bagheri et al., 2026
& PMC13380479 & EC & 7 \\

R015 & Karakasis et al., 2026
& PMC13391209 & EC & 6 \\

R016 & Chen et al., 2026
& PMC13506725 & EC & 29 \\

R017 & Nguyen et al., 2026
& PMC13446969 & EC & 15 \\

R018 & Zhou et al., 2026
& PMC13472815 & EC & 26 \\

R019 & Zhang et al., 2026
& PMC13473429 & EC & 7 \\

R020 & Wu et al., 2026
& PMC13486297 & EC & 10 \\

R021 & Mohamed et al., 2026
& PMC13495394 & EC & 5 \\

R022 & Bao et al., 2026
& PMC13507955 & EC & 10 \\

R023 & Wang et al., 2026
& PMC13522188 & EC & 4 \\

R024 & R. Murad et al., 2026
& PMC13287412 & EC & 4 \\

R025 & Na et al., 2025
& PMC12028552 & EC & 5 \\

R026 & Alsubaiei et al., 2025
& PMC12358186 & EC & 4 \\

R027 & Ding et al., 2025
& PMC12541855 & EC & 4 \\

R028 & AbuJwaid et al., 2025
& PMC12279844 & EC & 4 \\

R029 & Ng et al., 2026
& PMC13513794 & EC & 4 \\

R030 & Yin et al., 2026
& PMC13229865 & EC & 4 \\

R031 & Saifi et al., 2025
& PMC12494503 & EC & 5 \\

R032 & de Albuquerque Pessoa dos Santos et al., 2025
& PMC12374470 & EC & 5 \\

R033 & Duan et al., 2025
& PMC12312543 & EC & 5 \\

R034 & Nair et al., 2025
& PMC12289400 & EC & 5 \\

R035 & Ali et al., 2025
& PMC12594780 & EC & 5 \\

R036 & Khraisat et al., 2026
& PMC13591805 & EC & 5 \\

R037 & Romyasamit et al., 2026
& PMC13608589 & EC & 6 \\

R038 & Jiang et al., 2026
& PMC13611371 & EC & 6 \\

R039 & Manzoor et al., 2026
& PMC13700451 & EC & 6 \\

R040 & M. Manasrah et al., 2025
& PMC12778877 & EC & 6 \\

R041 & Ando et al., 2025
& PMC12346875 & EC & 4 \\

R042 & Theodorakopoulos et al., 2025
& PMC12623144 & EC & 5 \\

R043 & Yang et al., 2025
& PMC12542067 & EC & 4 \\

R044 & Cheung et al., 2026
& PMC12908938 & EC & 5 \\

R045 & Wang et al., 2026
& PMC12960011 & EC & 6 \\

R046 & Li et al., 2026
& PMC12992012 & EC & 6 \\

R047 & Castelo et al., 2026
& PMC12999590 & EC & 7 \\

R048 & AlJuma et al., 2026
& PMC13002440 & EC & 5 \\

R049 & Elfeky et al., 2026
& PMC13014514 & EC & 16 \\

R050 & Albeladi et al., 2026
& PMC13055524 & EC & 6 \\

R051 & Sbarra et al., 2026
& PMC13059020 & EC & 6 \\

R052 & Bano et al., 2026
& PMC13071367 & EC & 5 \\

R053 & Agrafiotis et al., 2026
& PMC13072659 & EC & 4 \\

R054 & Omar et al., 2026
& PMC13073525 & EC & 8 \\

R055 & Qazi et al., 2026
& PMC13073533 & EC & 3 \\

R056 & Zrineh et al., 2026
& PMC13094063 & EC & 7 \\

R057 & G{\l}uszak et al., 2026
& PMC13113669 & EC & 3 \\

R058 & Alhebshi et al., 2026
& PMC13136402 & EC & 6 \\

R059 & Ibrahim et al., 2026
& PMC13138499 & EC & 6 \\

R060 & Zaihan et al., 2026
& PMC13156516 & EC & 6 \\

R061 & Gricius et al., 2026
& PMC13163817 & EC & 8 \\

R062 & Din et al., 2026
& PMC13215653 & EC & 4 \\

R063 & Aldamen et al., 2026
& PMC13236339 & EC & 4 \\

R064 & Alasmary et al., 2026
& PMC13264147 & EC & 3 \\

R065 & Srisurapanont et al., 2026
& PMC13265653 & EC & 7 \\

R066 & Zeng et al., 2026
& PMC13272463 & EC & 22 \\

R067 & Sharma et al., 2025
& PMC13278781 & EC & 3 \\

R068 & Zhou et al., 2026
& PMC13286486 & EC & 5 \\

R069 & Chan et al., 2026
& PMC13292675 & EC & 4 \\

R070 & Al Ameer et al., 2026
& PMC13294293 & EC & 6 \\

R071 & Alnazari et al., 2026
& PMC13318564 & EC & 3 \\

R072 & Kanno et al., 2026
& PMC13328200 & EC & 12 \\

R073 & Li et al., 2026
& PMC13333436 & EC & 6 \\

R074 & Yang et al., 2025
& PMC13347903 & EC & 5 \\

R075 & Napoli et al., 2026
& PMC13399242 & EC & 4 \\

R076 & Darlison et al., 2026
& PMC13399252 & EC & 14 \\

R077 & de Farias et al., 2026
& PMC13399485 & EC & 3 \\

R078 & Sunaga et al., 2026
& PMC13402018 & EC & 6 \\

R079 & Kania et al., 2026
& PMC13413132 & EC & 16 \\

R080 & Khan et al., 2026
& PMC13413204 & EC & 7 \\

R081 & Delgado et al., 2026
& PMC13422579 & EC & 8 \\

R082 & Hafez et al., 2026
& PMC13427285 & EC & 6 \\

R083 & Tsai et al., 2026
& PMC13438215 & EC & 5 \\

R084 & Bardak et al., 2026
& PMC13439305 & EC & 6 \\

R085 & Kim et al., 2026
& PMC13459513 & EC & 6 \\

R086 & Murad et al., 2026
& PMC13465320 & EC & 3 \\

R087 & Pantaleao et al., 2026
& PMC13476533 & EC & 5 \\

R088 & Jiang et al., 2026
& PMC13486172 & EC & 9 \\

R089 & Cheema et al., 2026
& PMC13491151 & EC & 4 \\

R090 & Basavanna et al., 2026
& PMC13518020 & EC & 6 \\

R091 & Yin et al., 2026
& PMC13526583 & EC & 7 \\

R092 & Liu et al., 2026
& PMC13538021 & EC & 22 \\

R093 & Reyes et al., 2026
& PMC13541452 & EC & 14 \\

R094 & Han et al., 2026
& PMC13547470 & EC & 5 \\

R095 & Barboza et al., 2026
& PMC13554177 & EC & 8 \\

R096 & Bourisly et al., 2026
& PMC13565801 & EC & 4 \\

R097 & Barboza et al., 2026
& PMC13566465 & EC & 5 \\

R098 & Osama et al., 2026
& PMC13570013 & EC & 7 \\

R099 & Qiu et al., 2026
& PMC13577174 & EC & 14 \\

R100 & Schena et al., 2026
& PMC13582080 & EC & 7 \\

R101 & Hussain et al., 2026
& PMC13582391 & EC & 4 \\

R102 & Alshehri et al., 2026
& PMC13603284 & EC & 9 \\

\end{longtable}
\endgroup

\section{Result Alignment and the Full Evaluation Space}
\label{app:B}

\subsection{Study $\times$ Target Alignment}

Before DE evaluation, two researchers independently aligned the results delivered by each system to the predefined Study $\times$ Target grid. Alignment was based on study identity, comparison roles, outcome definitions, follow-up time points, populations, analysis conditions, and result types. The researchers aligned only the content already delivered by each system; they did not supplement missing system outputs with numbers from the original sources. Partial results, missing results, and unresolved ambiguities retained their original statuses and were scored uniformly by the program.

\subsection{DE Evaluation Space}

The DE evaluation covers 405 Study $\times$ Target units. The reference set contains 195 reported results: 171 binary results and 24 continuous results. Binary results contain event counts and denominators for both groups; continuous results contain means, standard deviations, and sample sizes for both groups. These yield a total of 828 numerical fields:
$171 \times 4 + 24 \times 6 = 828$.
A result is considered completely correct only when all four fields of a binary result or all six fields of a continuous result match the reference. Reported results that are not delivered in full remain in the fixed reference denominator. Complete candidates for ordinary non-positive targets and reference-conflict targets enter the candidate set but do not contribute to the correct-result numerator. Complete candidates must also satisfy the claim-closure requirements in Appendix~\ref{app:D2}.

\subsection{Evaluation Conditions}

DE results are reported for MetaRigor, Codex CLI, OpenHands CLI,
Manalyzer, meta-pipe, and w/o Task Contract.
The w/o Code Logic ablation does not apply to DE, and all
corresponding entries are marked N/A (not applicable).
EA configuration comparisons and mechanism ablations for the
other stages are described in Appendix~\ref{app:E}
under the same evaluation framework.

\section{Representative Task Contract and Baseline Prompt}
\label{app:C}

\subsection{Example DE Task Contract}

\textbf{Task object.} Determine whether an observation matches the specified Study $\times$ Target and whether its reported fields can be adopted for that target.

\textbf{Permitted facts.} The study, comparison, outcome, population, time point, and analysis conditions specified in the Protocol; original reports and Source Spans in the current Study Packet; fields already parsed from the current observation; and the predefined schema and numerical constraints.

\textbf{Prohibited facts.} Results from other Studies or Targets; results from other time points, populations, or comparison groups; model memory or inferences without supplied source support; and numbers absent from the current materials that are inferred backward from review conclusions.

\textbf{Delivery requirements.} Return match, mismatch, or unresolved, explain each field judgment, and provide a Source Anchor for every adopted field. Retain unresolved for insufficient evidence; do not fill in missing numerical values or modify deterministic fields generated by the program.

\subsection{Example CLI Baseline Prompt}

Complete the task using the supplied research question, Protocol, and study materials. Use only facts explicitly provided in the input files. Do not add studies, numbers, time points, comparison groups, or analysis populations. When evidence is insufficient, return unresolved and explain why. Follow the specified output format strictly, and do not combine results from other studies or time points.

This example illustrates only the core constraints of the baseline task; it does not replace the complete stage-specific prompts, schemas, and model configurations.

\section{Evaluation Metrics, Scoring Procedures, and Valid-Output Scope}
\label{app:D}

This appendix consolidates the metric names, evaluation levels, and supplementary results for Tables \ref{tab:comparative_performance} and \ref{tab:ablation_study} and the main-text MS scoring figure. Percentage metrics are rounded to two decimal places by default; reference-annotation agreement retains the one-decimal precision used in Table \ref{tab:app_agreement}; and O-MAE and descriptive risk ratios are rounded to three decimal places. N/A indicates that the configuration is not applicable
or that no corresponding result is available; it does
not represent a score of zero. 

\subsection{EA Metrics}

Precision, $F_1$, $F_2$, and $NNR$ are reported separately for title--abstract ($TA$) and full-text ($FT$) screening. Each stage uses micro-aggregation: the numbers of correctly included records ($TP$), system-predicted inclusions ($C$), and reference inclusions ($N$) are first summed across cases, and metrics are then calculated from these aggregate counts. Case-level scores are not calculated first and then averaged. The metrics are defined as follows
\begin{equation}
\mathrm{Precision} = \frac{TP}{C},
\qquad
F_1 = \frac{2\,TP}{N + C},
\qquad
F_2 = \frac{5\,TP}{4N + C},
\qquad
NNR = \frac{C}{TP}.    
\end{equation}
NNR is the number of system-included records per correctly included record and equals the reciprocal of precision. It expresses the screening burden in practical review settings and is not treated as performance evidence independent of precision.

When the denominator of any formula is zero, the evaluation script records the metric as null rather than zero. This treatment differs from N/A for inapplicable conditions and does not alter the micro-aggregation convention based on aggregate counts.

Study-material recovery is reported as the complete study-family recovery rate and the companion-report recovery rate, with reference denominators of 25 study families and 29 companion reports, respectively. The column labels Family Rec.\ and Comp.\ Rec.\ in the main-text tables correspond to these two rates. For the full configuration, 88.00\% corresponds to 22/25 study families, and 93.10\% corresponds to 27/29 companion reports. Screening and recovery results are presented in Tables \ref{tab:comparative_performance} and \ref{tab:ablation_study} of the main text for direct cross-system comparison.
\subsection{DE Metrics and Supplementary Results}
\label{app:D2}

\hl{Source validation is part of MetaRigor's DE stage. The extractor proposes
candidate items, and the DE verifier accepts an item only if every
numerical field it contains has a valid Source Anchor. If any field fails
validation, the entire item is rejected. Missing fields remain missing,
so a partial item can still be accepted if all its present fields pass
validation. The baseline systems do not implement an equivalent
Source Anchor-based source-binding mechanism; therefore, their delivered
DE outputs are scored directly without this validation step. To
distinguish extraction performance from the effect of this DE-specific
Claim Closure-only filtering mechanism, we additionally report
MetaRigor's results before filtering. The unfiltered outputs achieved
an exact-match Recall of 89.23\%, Precision of 83.25\%, Field Accuracy
of 91.79\%, and $F_1$ of 86.14\%.}

Let $N = 195$ be the number of reference results, $C$ the number of
scored items that are numerically complete and can be uniquely encoded,
and $E$ the number of those items whose numerical fields all match the
reference. A complete item contains all four required fields for a binary
outcome or all six for a continuous outcome. We compute:
\begin{equation}
\mathrm{Recall}=\frac{E}{N},
\qquad
\mathrm{Precision}=\frac{E}{C},
\qquad
F_1=\frac{2E}{N+C}.
\end{equation}

Let $M$ be the number of scored numerical fields that match their
reference values, including correct fields in partial items. Field
accuracy uses the fixed denominator of 828 reference fields:
\begin{equation}
\mathrm{Field\ Acc.} = \frac{M}{828}.
\label{eq:field_acc}
\end{equation}
The corresponding DE performance across systems and configurations
is summarized in Table~\ref{tab:d1}.

\begin{table}[ht]
  \caption{DE performance at the complete result and field levels (\%)}
  \label{tab:d1}
  \centering
  \begingroup
  \small
  \color{black}
  \setlength{\tabcolsep}{2pt}
  \setlength{\aboverulesep}{1.5pt}
  \setlength{\belowrulesep}{1.5pt}
  \renewcommand{\arraystretch}{1.15}
  \begin{tabular*}{\linewidth}{@{\extracolsep{\fill}}lcccc}
    \toprule
    \textbf{System / configuration}
    & \textbf{Complete result recall}
    & \textbf{Precision}
    & \textbf{$\text{F}_1$}
    & \textbf{Field accuracy} \\
        \midrule
    Codex CLI
    & 84.62 & 84.62 & 84.62 & 85.99 \\
    OpenHands CLI
    & 71.79 & 78.21 & 74.87 & 77.66 \\
    Manalyzer
    & 32.31 & 98.44 & 48.65 & 35.75 \\
    meta-pipe
    & 40.51 & 76.70 & 53.02 & 44.93 \\
    MetaRigor
    & 87.18 & 85.43 & 86.29 & 88.89 \\
    w/o Task Contract
    & 83.08 & 80.20 & 81.61 & 86.84 \\
    w/o Code Logic
    & N/A & N/A & N/A & N/A \\
    \bottomrule
  \end{tabular*}
  \endgroup
\end{table}

\subsection{RoB Metrics}

RoB reports domain agreement, overall-judgment agreement, and high risk recall, corresponding to Domain, Judgment, and High Risk in the main-text tables. The denominators are 160 domain judgments for domain agreement, 32 study–result overall judgments for overall-judgment agreement, and 13 objects labeled as overall high risk in the reference for high risk recall. Missing or invalid judgments are not counted as correct matches and are not removed from the fixed denominators.

The current program uses a simplified overall aggregation rule: if any domain is High risk, the overall judgment is High risk; if all domains are Low risk, the overall judgment is Low risk; otherwise, it is Some concerns. Context-dependent judgments in which multiple Some concerns domains jointly warrant an overall High risk judgment are not yet included in the automated aggregation. Overall errors may therefore arise both from domain judgments and from the coverage of the aggregation rules. Results for each configuration are presented in Tables \ref{tab:comparative_performance} and \ref{tab:ablation_study} of the main text.

\subsection{EC Metrics and Valid-Output Scope}

EC reports domain agreement, overall downgrade-count agreement, and mean absolute error in the overall downgrade count, corresponding to Domain, Downgrade, and MAE in the main-text tables. Domain agreement uses 500 domain judgments as the denominator under exact matching, with boundary labels treated as distinct valid categories for exact matching. Overall downgrade-count agreement uses all 100 analyses as the denominator and compares only explicitly reported overall downgrade counts. Missing or invalid outputs count as non-matches and are not imputed from the final certainty level or the sum of domain downgrades.

The interval constraint used in overall certainty synthesis is based on
the mapping from domain labels to permitted downgrade intervals shown in
Table~\ref{tab:ec_interval_mapping}.

\begin{table}[ht]
  \caption{Mapping from EC domain labels to permitted downgrade intervals.}
  \label{tab:ec_interval_mapping}
  \centering
  \begingroup
  \small
  \color{black}
  \setlength{\tabcolsep}{2pt}
  \setlength{\aboverulesep}{1.5pt}
  \setlength{\belowrulesep}{1.5pt}
  \renewcommand{\arraystretch}{1.15}
  \begin{tabular*}{\linewidth}{@{\extracolsep{\fill}}lc}
    \toprule
    \textbf{Domain label}
    & \textbf{Permitted downgrade interval}
   \\
    \midrule
    NOT\_SERIOUS    &$[0,0]$\\
    BORDERLINE\_NOT\_SERIOUS\     &$[0, 1]$\\
    SERIOUS    &$[1, 1]$\\
    BORDERLINE\_SERIOUS\   &$[1, 2]$\\
    VERY\_SERIOUS    &$[2, 2]$ \\
    \bottomrule
  \end{tabular*}
  \endgroup
\end{table}

The same mapping is applied to all five EC domains. The overall lower bound is the maximum of the five domain-level lower bounds, whereas the overall upper bound is the sum of the five domain-level upper bounds, capped at three. Borderline labels preserve uncertainty between two adjacent downgrade levels; they do not determine the final overall downgrade by themselves.

The interval constraint limits the admissible numerical range of the final downgrade count, but it does not determine whether domain-level limitations are independent, overlapping, or caused by the same underlying issue. Resolving potential duplicate downgrading remains a semantic synthesis decision performed by the EC synthesis model. We therefore do not report a separate synthesis-only accuracy estimate for this step within the current evaluation framework.

The mean absolute error in the overall downgrade count (O-MAE, corresponding to MAE in the main text) is calculated only over the paired set V in which both the system and the reference provide valid overall downgrade counts:
\begin{equation}
O\text{-MAE} = \frac{1}{|V|} \sum_{i\in V} \big|\hat{\Delta}_i - \Delta_i\big|.
\label{eq:omae}
\end{equation}
where $\Delta_i$ and $\hat{\Delta}_i$ denote the reference
and system overall downgrade counts, respectively.
O-MAE is calculated over cases with valid overall downgrade
counts from both the system and the reference.
Overall downgrade-count agreement uses the fixed denominator
of 100 analyses and counts missing or invalid outputs as
non-matches.

\subsection{MS Scoring and Complete Dimension-Level Results}

\subsubsection{Scoring Criteria and Review Procedure}

MS uses seven scoring criteria, each rated on a 0–4 scale, with weights listed in Table \ref{tab:d3}. The scoring anchors are: 4, fully satisfies the criterion with no substantive issues; 3, largely satisfies the criterion with localized issues; 2, partially satisfies the criterion but contains substantive omissions requiring manual correction; 1, fails to satisfy most requirements and may mislead readers; and 0, a critical failure, fabricated facts, or a major deviation from the Protocol.

\begin{table}[ht]
  \caption{MS scoring dimensions, figure labels, and weights}
  \label{tab:d3}
  \centering
  \begingroup
  \small
  \color{black}
  \setlength{\tabcolsep}{2pt}
  \setlength{\aboverulesep}{1.5pt}
  \setlength{\belowrulesep}{1.5pt}
  \renewcommand{\arraystretch}{1.15}
  \begin{tabular*}{\linewidth}{@{\extracolsep{\fill}}lcc}
    \toprule
    \textbf{Dimension}
    & \textbf{Figure label}
    & \textbf{Weight} \\
    \midrule
    Factual fidelity                & Facts         & 25\% \\
    Protocol and execution fidelity & Protocol      & 15\% \\
    Material coverage               & Coverage      & 15\% \\
    Interpretive calibration        & Calibration   & 15\% \\
    Source traceability             & Traceability  & 10\% \\
    Cross-section consistency       & Consistency   & 10\% \\
    Scientific organization         & Organization  & 10\% \\
    \bottomrule
  \end{tabular*}
  \endgroup
\end{table}

DeepSeek, Luna, and the human reviewers assessed manuscripts from the same ten cases, with mutually independent scoring contexts. Both human reviewers had research experience in biomedical artificial intelligence, medical image analysis, and meta-analysis. They scored independently using the same materials, weights, and scoring anchors, without viewing each other’s scores, and did not participate in system execution, manuscript generation, or evaluation of the other stages. Human review used anonymous configuration labels and randomized configuration order; system, model, and experimental-condition identities were not disclosed to the reviewers.

For each manuscript and dimension, the initial score was the arithmetic mean of the two reviewers’ scores, rounded to the prespecified precision. When the difference between their scores exceeded 0.4 points on any dimension, they discussed the scoring rationale for that dimension only and adopted a mutually agreed final score. Discussion did not alter the manuscript or scores on other dimensions. The number of reviewers, criteria, weights, disagreement threshold, and adjudication rules were fixed before the formal evaluation to ensure procedural consistency.

\subsubsection{Evaluator-Specific and Composite Scoring}

Let $r_{srjk}$ denote the final score for manuscript j from system s, assigned by evaluator category r on dimension k, and let $w_k$ denote the dimension weight. The manuscript score on a 100-point scale and the system-level score for that evaluator category are, respectively:
\begin{equation}
\mathrm{Score}_{s,r,j} = 100 \sum_{k=1}^{7} w_k \frac{r_{s,r,j,k}}{4},\quad \overline{\mathrm{Score}}_{s,r} = \frac{1}{10} \sum_{j=1}^{10} \mathrm{Score}_{s,r,j}.
\end{equation}
Weights are expressed as proportions; for example, 25\% is entered as 0.25. The evaluator categories r are DeepSeek, Luna, and human review. The human category uses the final scores aggregated from the two reviewers as described above; the two human reviewers are not treated as two independent evaluator categories. Tables~\ref{tab:ms_deepseek}, \ref{tab:ms_luna}, \ref{tab:ms_human}, and \ref{tab:ms_overall} provide the complete dimension-level results. Each dimension is scored from 0 to 4, and total scores are expressed on a 100-point scale.

The MS scoring figure in the main text reports a cross-evaluator composite score as Overall, calculated as the equally weighted mean of the human, Luna, and DeepSeek evaluation results:
\begin{equation}
\mathrm{Overall}_s \;=\;
\frac{
\overline{\mathrm{Score}}_{s,\mathrm{Human}}
+
\overline{\mathrm{Score}}_{s,\mathrm{Luna}}
+
\overline{\mathrm{Score}}_{s,\mathrm{DeepSeek}}
}{3}.
\end{equation}
System-level scores for each evaluator category and Overall are aggregated from unrounded scores and rounded to two decimal places only for final presentation. Overall is 89.63 for MetaRigor and 71.63 for w/o Code Logic. The composite score does not replace evaluator-specific results; the three category-specific totals and the composite score are reported side by side in Table~\ref{tab:ms_overall}.


\begin{table}[ht]
\centering
\scriptsize
\setlength{\tabcolsep}{1pt}
\setlength{\aboverulesep}{1pt}
\setlength{\belowrulesep}{1pt}
\renewcommand{\arraystretch}{1.0}

\caption[Manuscript evaluation by DeepSeek.]{%
\protect\hl{Manuscript evaluation by DeepSeek.}%
}

\label{tab:ms_deepseek}

\begin{tabular*}{\linewidth}{@{\extracolsep{\fill}}lcccccccc}
\toprule
\textbf{System}
& \textbf{Factual}
& \textbf{Protocol}
& \textbf{Material}
& \textbf{Interpretive}
& \textbf{Source}
& \textbf{Cross-section}
& \textbf{Scientific}
& \textbf{Total} \\
\textbf{/configuration}
& \textbf{fidelity}
& \textbf{execution fidelity}
& \textbf{coverage}
& \textbf{calibration}
& \textbf{traceability}
& \textbf{consistency}
& \textbf{organization}
& \\
\midrule
\hl{meta-pipe}
& 4.00 & 2.60 & 4.00 & 3.80
& 2.80 & 3.40 & 2.70 & 86.25 \\
Codex CLI
& 2.80 & 2.10 & 2.90 & 2.90 & 2.10 & 2.80 & 2.90 & 66.63 \\
OpenHands CLI
& 3.30 & 2.20 & 3.40 & 3.70 & 2.10 & 3.20 & 2.30 & 74.50 \\
MetaRigor
& 4.00 & 3.90 & 3.50 & 3.70 & 3.90 & 3.60 & 3.70 & 94.63 \\
w/o Task Contract
& 3.80 & 2.30 & 3.50 & 3.50 & 2.30 & 3.40 & 2.80 & 79.88 \\
w/o Code Logic
& 3.70 & 2.50 & 3.70 & 3.40 & 2.50 & 3.80 & 2.30 & 80.63 \\
\bottomrule
\end{tabular*}

\end{table}


\begin{table}[ht]
\centering
\scriptsize
\setlength{\tabcolsep}{1pt}
\setlength{\aboverulesep}{1pt}
\setlength{\belowrulesep}{1pt}
\renewcommand{\arraystretch}{1.0}

\caption[Manuscript evaluation by Luna.]{%
\protect\hl{Manuscript evaluation by Luna.}%
}
\label{tab:ms_luna}

\begin{tabular*}{\linewidth}{@{\extracolsep{\fill}}lcccccccc}
\toprule
\textbf{System}
& \textbf{Factual}
& \textbf{Protocol}
& \textbf{Material}
& \textbf{Interpretive}
& \textbf{Source}
& \textbf{Cross-section}
& \textbf{Scientific}
& \textbf{Total} \\
\textbf{/configuration}
& \textbf{fidelity}
& \textbf{execution fidelity}
& \textbf{coverage}
& \textbf{calibration}
& \textbf{traceability}
& \textbf{consistency}
& \textbf{organization}
& \\
\midrule
\hl{meta-pipe}
& 2.50 & 2.20 & 2.90 & 2.90
& 2.90 & 2.70 & 2.60 & 66.13 \\
Codex CLI
& 1.90 & 1.70 & 2.10 & 2.40 & 1.90 & 1.80 & 2.60 & 50.88 \\
OpenHands CLI
& 3.10 & 2.00 & 2.90 & 3.40 & 1.70 & 2.70 & 2.50 & 67.75 \\
MetaRigor
& 3.90 & 2.60 & 3.80 & 3.90 & 3.40 & 3.50 & 3.20 & 88.25 \\
w/o Task Contract
& 3.10 & 1.90 & 2.80 & 2.60 & 2.00 & 2.50 & 2.40 & 64.00 \\
w/o Code Logic
& 2.80 & 2.00 & 3.20 & 3.00 & 2.00 & 2.40 & 1.90 & 64.00 \\
\bottomrule
\end{tabular*}

\end{table}


\begin{table}[ht]
\centering
\scriptsize
\setlength{\tabcolsep}{1pt}
\setlength{\aboverulesep}{1pt}
\setlength{\belowrulesep}{1pt}
\renewcommand{\arraystretch}{1.0}

\caption[Manuscript evaluation by human reviewers.]{%
\protect\hl{Manuscript evaluation by human reviewers.}%
}
\label{tab:ms_human}

\begin{tabular*}{\linewidth}{@{\extracolsep{\fill}}lcccccccc}
\toprule
\textbf{System}
& \textbf{Factual}
& \textbf{Protocol}
& \textbf{Material}
& \textbf{Interpretive}
& \textbf{Source}
& \textbf{Cross-section}
& \textbf{Scientific}
& \textbf{Total} \\
\textbf{/configuration}
& \textbf{fidelity}
& \textbf{execution fidelity}
& \textbf{coverage}
& \textbf{calibration}
& \textbf{traceability}
& \textbf{consistency}
& \textbf{organization}
& \\
\midrule
\hl{meta-pipe}
& 2.70 & 2.60 & 2.80 & 2.60
& 2.50 & 2.60 & 3.40 & 68.13 \\
Codex CLI
& 2.10 & 2.30 & 2.40 & 2.60 & 2.10 & 1.90 & 2.70 & 57.25 \\
OpenHands CLI
& 3.20 & 3.20 & 3.10 & 3.50 & 1.90 & 2.70 & 3.70 & 77.50 \\
MetaRigor
& 3.80 & 3.30 & 3.80 & 3.70 & 2.70 & 3.30 & 2.70 & 86.00 \\
w/o Task Contract
& 2.90 & 2.90 & 3.30 & 2.90 & 2.10 & 2.60 & 3.20 & 72.00 \\
w/o Code Logic
& 3.00 & 2.50 & 3.40 & 3.50 & 2.10 & 2.30 & 2.10 & 70.25 \\
\bottomrule
\end{tabular*}

\end{table}


\begin{table}[ht]
\centering
\scriptsize
\setlength{\tabcolsep}{3pt}
\setlength{\aboverulesep}{1pt}
\setlength{\belowrulesep}{1pt}
\renewcommand{\arraystretch}{1.0}

\caption[Evaluator-specific totals and cross-evaluator composite scores.]{%
\protect\hl{Evaluator-specific totals and cross-evaluator composite scores.}%
}
\label{tab:ms_overall}

\begin{tabular*}{\linewidth}{@{\extracolsep{\fill}}lcccc}
\toprule
\textbf{System / configuration}
& \textbf{DeepSeek}
& \textbf{Luna}
& \textbf{Human review}
& \textbf{Overall} \\
\midrule
\hl{meta-pipe}
& 86.25 & 66.13 & 68.13 & 73.50 \\
Codex CLI
& 66.63 & 50.88 & 57.25 & 58.25 \\
OpenHands CLI
& 74.50 & 67.75 & 77.50 & 73.25 \\
MetaRigor
& 94.63 & 88.25 & 86.00 & 89.63 \\
w/o Task Contract
& 79.88 & 64.00 & 72.00 & 71.96 \\
w/o Code Logic
& 80.63 & 64.00 & 70.25 & 71.63 \\
\bottomrule
\end{tabular*}

\end{table}

\subsection{Claim Closure and Reference Recall}

Source verifiability is evaluated using two complementary metrics: Claim Closure and Reference Recall, which capture technical binding validity and evidence recovery, respectively. Claim Closure measures whether a delivered output is attached to a permitted and retrievable source location. Reference Recall measures whether the system recovers manually identified evidence spans that support the corresponding output. These metrics use different evaluation units and denominators and are therefore reported separately. Neither metric establishes that the retrieved text semantically supports the corresponding judgment, nor do they replace task-level correctness metrics. The corresponding source-verifiability results are summarized in Table~\ref{tab:d6}.

\begin{table}[ht]
\caption{Source-verifiability metrics under internal ablations (\%).}
\label{tab:d6}
\centering
\scriptsize
\setlength{\tabcolsep}{2pt}
\setlength{\aboverulesep}{1.5pt}
\setlength{\belowrulesep}{1.5pt}
\renewcommand{\arraystretch}{1.2}

\begin{tabular*}{\linewidth}{@{\extracolsep{\fill}}lcccccc@{}}
\toprule

\textbf{Configuration}
& \multicolumn{2}{c}{\textbf{DE}}
& \multicolumn{2}{c}{\textbf{RoB}}
& \multicolumn{2}{c}{\textbf{EC}}
\\

\cmidrule(lr){2-3}
\cmidrule(lr){4-5}
\cmidrule(lr){6-7}

&
\shortstack{\textbf{Claim}\\\textbf{Closure} $\uparrow$}
&
\shortstack{\textbf{Reference}\\\textbf{Recall} $\uparrow$}
&
\shortstack{\textbf{Claim}\\\textbf{Closure} $\uparrow$}
&
\shortstack{\textbf{Reference}\\\textbf{Recall} $\uparrow$}
&
\shortstack{\textbf{Claim}\\\textbf{Closure} $\uparrow$}
&
\shortstack{\textbf{Reference}\\\textbf{Recall} $\uparrow$}
\\

\midrule

MetaRigor
& 95.70
& 88.21
& 74.67
& 50.85
& 100.00
& 82.98
\\

\midrule

w/o Task Contract
& 95.38
& 86.67
& 76.68
& 49.15
& 68.07
& 69.50
\\

\midrule

w/o Code Logic
& N/A
& N/A
& 76.84
& 52.78
& 97.63
& 90.78
\\

\bottomrule
\end{tabular*}

\end{table}

\subsubsection{Claim Closure}

Let n denote the number of delivered units submitted to source verification and v the number with valid source attachments. Claim Closure is defined as:
\begin{equation}
\text{Claim Closure} = \frac{v}{n}.
\end{equation}
For DE, the verification unit is a delivered numerical field in the unfiltered system output. MetaRigor achieves 891/931 = 95.70\%, while w/o Task Contract achieves 908/952 = 95.38\%. Undelivered fields are not included in this denominator. This evaluation precedes the Item-level closure filtering used for the task metrics in Appendix~\ref{app:D2}, where an Item is retained only when all delivered numerical fields have valid source attachments. Consequently, the retained DE evaluation set has 100\% closure by construction in this analysis.

RoB applies the same source-attachment validity criterion as DE to unfiltered delivered RoB evidence. EC Claim Closure likewise measures the technical binding of delivered evidence to valid source attachments before task-level judgment scoring.

\subsubsection{Reference Recall}

Let H denote the number of matched reference evidence units and N\_ref the number of predefined reference units. Reference Recall is defined as:
\begin{equation}
\text{Reference Recall}=\frac{H}{N_{\mathrm{ref}}}.
\end{equation}
For DE, the denominator is 195 reference reported-result Items. An Item is counted once if an accepted reference span is recovered; repeated matches do not increase the numerator. MetaRigor recovers 172/195 = 88.21\%, and w/o Task Contract recovers 169/195 = 86.67\%. This metric assesses recovery of reference evidence and differs from complete result recall in Appendix~\ref{app:D2}.

For RoB, Reference Recall is evaluated against 413 reference signaling questions (SQs), with repeated citations to the same reference unit counted only once. This reference denominator is distinct from the 704 task signaling questions reported in Appendix \ref{app:A}. For EC, the reference set contains 141 evidence links after deduplication by case, domain, and evidence ID; each deduplicated reference unit contributes at most once to the numerator.

Claim Closure, Reference Recall, and task correctness need not change in the same direction. For example, w/o Code Logic obtains higher Reference Recall than the full configuration in both RoB and EC, while its task-level judgment performance is lower. Source-verifiability metrics are therefore reported as complementary diagnostics rather than substitutes for task accuracy.

\section{Execution Configurations, Baselines, and Ablation Conditions}
\label{app:E}

\subsection{Stage-Controlled Comparisons}

Sections~\ref{sec:comparative}--\ref{sec:ablation} of the main text present system-level comparisons using the same research questions, permitted materials, fixed upstream inputs, and task requirements. DE, RoB, EC, and MS all use deepseek-v4-flash; EA retains each system’s formal execution configuration. Raw outputs, tool returns, and failure states were retained
for traceability. 

\hl{
All five systems were evaluated using the same research
questions, Protocols, task-visible input materials, task
definitions, evaluation reference sets, and tool-access
permissions. Network search was disabled for all systems
to prevent direct retrieval of the original literature used
as the evaluation basis through external search.
The same execution budget of 1,800 seconds was allocated
to each system at each stage. Evaluation reference answers
were not provided as task inputs and were used exclusively
for scoring.

The systems differed in their native error-correction
capabilities. Neither MetaRigor nor meta-pipe included
a dedicated error-correction stage. Codex CLI and
OpenHands CLI could leverage iterative agent execution
within the same session to further inspect, analyze, and
revise previously delivered outputs, whereas Manalyzer
included an explicit correction stage. In contrast,
when programmatic verification identified errors or
unresolved issues, MetaRigor retained the corresponding
failure or unresolved status without automatically
triggering additional model calls for retry or revision.
These native differences were preserved in the comparison
rather than artificially equalized, as the evaluation
aimed to compare the systems' respective workflow
capabilities under matched task and resource conditions.
}

Codex CLI and OpenHands CLI retain their native execution interfaces and tool-orchestration mechanisms and serve as system-level agent baselines, rather than isolated model or prompt baselines. The comparison concerns workflow organization, including result identity, task contracts, programmatic rules, source binding, and delivery requirements. Because EA uses each system’s formal configuration, its comparison includes the associated configuration differences; between-system differences are not attributed exclusively to any one component. Cross-stage artifact analyses are interpreted separately within the scope specified in Appendix \ref{app:G}.

In the full EA configuration, DeepSeek-V4-Flash and GPT-5.6 Luna execute the two independent screening paths, respectively. GPT-5.6 Terra is used to adjudicate disagreements at the full-text stage. The two model-based evaluators for MS are DeepSeek and Luna. The recorded software versions are Codex CLI 0.142.4 and OpenHands CLI 1.16.0. Specific roles are listed in Table \ref{tab:e1}.

\begin{table}[ht]
  \caption{Model roles and CLI versions}
  \label{tab:e1}
  \centering
  \begingroup
  \small
  \setlength{\tabcolsep}{2pt}
  \setlength{\aboverulesep}{1.5pt}
  \setlength{\belowrulesep}{1.5pt}
  \renewcommand{\arraystretch}{1.15}
  \begin{tabular*}{\linewidth}{@{\extracolsep{\fill}}l l}
    \toprule
    \textbf{Task / role} & \textbf{Model or software identifier} \\
    \midrule
    Generation model for DE, RoB, EC, and MS & deepseek-v4-flash \\
    EA independent screening path 1 & deepseek-v4-flash \\
    EA independent screening path 2 & gpt-5.6-luna \\
    EA full-text disagreement adjudication & gpt-5.6-terra \\
    MS evaluator (DeepSeek) & deepseek-v4-flash \\
MS evaluator (Luna) & gpt-5.6-luna \\
    Codex CLI & deepseek-v4-flash, 0.142.4 \\
        OpenHands CLI & deepseek-v4-flash, 1.16.0 series \\
    \hl{Manalyzer} & deepseek-v4-flash, commit 18544a4 \\
\hl{meta-pipe} & deepseek-v4-flash, commit 5c5c3f0 \\
    \bottomrule
  \end{tabular*}
  \endgroup
\end{table}

\subsection{Configuration Comparisons and Applicable Stages}

The full EA configuration uses DeepSeek-V4-Flash and GPT-5.6 Luna for the two independent screening paths, respectively, with GPT-5.6 Terra adjudicating disagreements at the full-text stage. We compare this configuration with Dual Luna and Dual DeepSeek. Dual Luna uses GPT-5.6 Luna for both screening paths, whereas Dual DeepSeek uses DeepSeek-V4-Flash for both screening paths.

The applicable stages and configuration changes are summarized in Table~\ref{tab:e2}.
\begin{table}[ht]

\caption[Comparison conditions and scope of applicability]{%
\protect\hl{Comparison conditions and scope of applicability}%
}

  \label{tab:e2}
  \centering
  \begingroup
  \small
  \setlength{\tabcolsep}{3pt}
  \setlength{\aboverulesep}{1.5pt}
  \setlength{\belowrulesep}{1.5pt}
  \renewcommand{\arraystretch}{1.2}
  \begin{tabular*}{\linewidth}{@{\extracolsep{\fill}}l l p{5.5cm}}
    \toprule
    \textbf{Condition} & \textbf{Applicable stages} & \textbf{Core configuration or change} \\
    \midrule
    MetaRigor & EA, DE, RoB, EC, MS & Full workflow: result identity, Task Contract, programmatic rules, and source binding \\
    Codex CLI & EA, DE, RoB, EC, MS & Native CLI agent execution \\
    OpenHands CLI & EA, DE, RoB, EC, MS & Native agent workflow \\
    \hl{Dual Luna} & EA & GPT-5.6 Luna covers all evaluation roles
in this EA comparison, including full-text adjudication \\
    Dual DeepSeek & EA & DeepSeek covers all evaluation roles in this EA comparison, including full-text adjudication \\
    w/o Task Contract & DE, RoB, EC, MS & Explicit Task Contract removed \\
    w/o Code Logic & RoB, EC, MS & Selected formalized rules delegated to model judgment; not applicable to DE \\
    \bottomrule
  \end{tabular*}
  \endgroup
\end{table}

\section{Overwriting of Tool Outputs During Subsequent Generation}
\label{app:F}

\hl{
This section examines the calculator calls recorded in the w/o Task Contract execution trace. The model autonomously initiated 8 calculator calls. These constituted all calculator calls in that run, rather than a post hoc sample, and originated from 8 distinct cases. For each call, the calculator input, calculator output, and corresponding final output were retained for analysis.

The input arguments for all 8 calls were consistent with the frozen upstream inputs. All labels returned by the calculator matched the reference and were included in subsequent model inputs, but 4 were overwritten during final generation with labels that did not match the reference. For example, in the short-term stroke rehospitalization case, SERIOUS was changed to $\text{NOT\_SERIOUS}$ solely because the confidence interval did not cross the null value, while the information-size criterion was omitted. The proportion 4/8 describes overwriting among the observed calculator calls and is not an overall error rate for the EC evaluation.
}

\section{Detailed Results of the Full Workflow Comparison and Analysis of Error Propagation}
\label{app:G}

\subsection{Experimental Setup}

We compared MetaRigor, meta-pipe, Codex CLI, and OpenHands CLI across seven cases. The end-to-end evaluation began with data extraction, abbreviated as DE, rather than evidence acquisition, abbreviated as EA, to maintain a consistent evaluation set across systems. Including EA could result in system-specific study sets due to differences in screening decisions. False-positive inclusions and false-negative exclusions would alter the evidence passed to downstream stages, complicating the construction of a shared reference benchmark and making it difficult to distinguish errors originating from screening from those introduced during subsequent stages. We therefore evaluated EA separately using its stage-specific benchmark. In the four-stage end-to-end evaluation, the outputs of each stage were passed sequentially to downstream stages. DE outputs were supplied to risk of bias assessment, abbreviated as RoB, and statistical synthesis. The outputs of these two processes were jointly passed to evidence certainty assessment, abbreviated as EC, followed by manuscript writing, abbreviated as MS. Each downstream stage used the actual upstream artifacts delivered by the corresponding system.

The evaluation covered 2,133 reference facts in the final
manuscript package, comprising 869 for DE, 906 for RoB,
114 for EC, and 244 for MS. DE, RoB, and EC facts were
evaluated in their corresponding artifacts included in
the package, while MS facts were evaluated in the
generated manuscript. Precision was defined as the proportion of expressed items that were correct, recall as the proportion of reference items correctly reported, and $F_1$ as their harmonic mean. For each stage and each case, counts were aggregated before the corresponding metrics were calculated. Incorrectly reported items were included in the expressed item count, whereas omitted items remained in the reference denominator. 

\subsection{Stage Wise and Case Specific Results}

Table~\ref{tab:g1} presents the results of all four systems across
individual stages. MetaRigor achieved $F_1$ scores of 86.89\%,
63.66\%, 64.00\%, and 92.37\% in DE, RoB, EC, and MS, respectively.
Its performance in DE was comparable to that of Codex CLI,
whereas the differences were more pronounced in RoB, EC, and MS.
meta-pipe achieved relatively high coverage in EC, but substantial
coverage gaps remained in DE and RoB.

\begin{table}[htbp]
\centering
\caption[Stage-wise comparison across seven cases.]{%
\protect\hl{Stage-wise comparison across seven cases.}%
}
\label{tab:g1}

\begingroup
\small
\setlength{\tabcolsep}{4pt}
\renewcommand{\arraystretch}{1.15}

\begin{tabularx}{\linewidth}{
  @{}cl*{6}{>{\centering\arraybackslash}X}@{}
}
\toprule
\textbf{Stage}
& \textbf{System}
& \textbf{Correct}
& \textbf{Expressed}
& \textbf{Reference}
& \textbf{Precision}
& \textbf{Recall}
& \textbf{F\textsubscript{1}} \\
\midrule

\multirow{4}{*}{DE}
& MetaRigor
& 736 & 825 & 869
& 89.21\% & 84.70\% & 86.89\% \\

& meta-pipe
& 325 & 381 & 869
& 85.30\% & 37.40\% & 52.00\% \\

& Codex CLI
& 725 & 822 & 869
& 88.20\% & 83.43\% & 85.75\% \\

& OpenHands CLI
& 681 & 794 & 869
& 85.77\% & 78.37\% & 81.90\% \\
\midrule

\multirow{4}{*}{RoB}
& MetaRigor
& 515 & 712 & 906
& 72.33\% & 56.84\% & 63.66\% \\

& meta-pipe
& 244 & 516 & 906
& 47.29\% & 26.93\% & 34.32\% \\

& Codex CLI
& 430 & 615 & 906
& 69.92\% & 47.46\% & 56.54\% \\

& OpenHands CLI
& 147 & 540 & 906
& 27.22\% & 16.23\% & 20.33\% \\
\midrule

\multirow{4}{*}{EC}
& MetaRigor
& 64 & 86 & 114
& 74.42\% & 56.14\% & 64.00\% \\

& meta-pipe
& 72 & 114 & 114
& 63.16\% & 63.16\% & 63.16\% \\

& Codex CLI
& 45 & 110 & 114
& 40.91\% & 39.47\% & 40.18\% \\

& OpenHands CLI
& 43 & 92 & 114
& 46.74\% & 37.72\% & 41.75\% \\
\midrule

\multirow{4}{*}{MS}
& MetaRigor
& 224 & 241 & 244
& 92.95\% & 91.80\% & 92.37\% \\

& meta-pipe
& 173 & 224 & 244
& 77.23\% & 70.90\% & 73.93\% \\

& Codex CLI
& 110 & 217 & 244
& 50.69\% & 45.08\% & 47.72\% \\

& OpenHands CLI
& 99 & 184 & 244
& 53.80\% & 40.57\% & 46.26\% \\

\bottomrule
\end{tabularx}

\endgroup
\end{table}

Table~\ref{tab:g2} presents the results for each of the seven cases,
while Table~\ref{tab:g3} summarizes the overall results and provides
an additional comparison excluding MS.
MetaRigor achieved an overall $F_1$ of 77.01\%.
Even after excluding manuscript writing facts, its $F_1$ remained
at 74.89\%, outperforming all other systems.


\begin{table}[htbp]
\centering
\caption[Results for each of the seven cases across DE, RoB, EC, and MS.]{%
\protect\hl{Results for each of the seven cases across DE, RoB, EC, and MS.}%
}
\label{tab:g2}

\begingroup
\small
\setlength{\tabcolsep}{3pt}
\renewcommand{\arraystretch}{1.15}

\begin{tabularx}{\linewidth}{
  @{}ll*{6}{>{\centering\arraybackslash}X}@{}
}
\toprule
\textbf{Case}
& \textbf{System}
& \textbf{Correct}
& \textbf{Expressed}
& \textbf{Reference}
& \textbf{Precision}
& \textbf{Recall}
& \textbf{F\textsubscript{1}} \\
\midrule

\multirow{4}{*}{lin-2025}
& MetaRigor
& 309 & 386 & 485
& 80.05\% & 63.71\% & 70.95\% \\

& meta-pipe
& 237 & 374 & 485
& 63.37\% & 48.87\% & 55.18\% \\

& Codex CLI
& 285 & 390 & 485
& 73.08\% & 58.76\% & 65.14\% \\

& OpenHands CLI
& 143 & 356 & 485
& 40.17\% & 29.48\% & 34.01\% \\
\midrule

\multirow{4}{*}{\shortstack[l]{turalde-mapili-\\2023}}
& MetaRigor
& 111 & 119 & 119
& 93.28\% & 93.28\% & 93.28\% \\

& meta-pipe
& 73 & 99 & 119
& 73.74\% & 61.34\% & 66.97\% \\

& Codex CLI
& 98 & 119 & 119
& 82.35\% & 82.35\% & 82.35\% \\

& OpenHands CLI
& 67 & 106 & 119
& 63.21\% & 56.30\% & 59.56\% \\
\midrule

\multirow{4}{*}{zhong-2022}
& MetaRigor
& 97 & 131 & 129
& 74.05\% & 75.19\% & 74.62\% \\

& meta-pipe
& 55 & 71 & 129
& 77.46\% & 42.64\% & 55.00\% \\

& Codex CLI
& 84 & 127 & 129
& 66.14\% & 65.12\% & 65.62\% \\

& OpenHands CLI
& 84 & 124 & 129
& 67.74\% & 65.12\% & 66.40\% \\
\midrule

\multirow{4}{*}{balbaa-2025}
& MetaRigor
& 317 & 394 & 419
& 80.46\% & 75.66\% & 77.98\% \\

& meta-pipe
& 128 & 209 & 419
& 61.24\% & 30.55\% & 40.76\% \\

& Codex CLI
& 174 & 236 & 419
& 73.73\% & 41.53\% & 53.13\% \\

& OpenHands CLI
& 163 & 212 & 419
& 76.89\% & 38.90\% & 51.66\% \\
\midrule

\multirow{4}{*}{meng-2025}
& MetaRigor
& 177 & 220 & 243
& 80.45\% & 72.84\% & 76.46\% \\

& meta-pipe
& 137 & 183 & 243
& 74.86\% & 56.38\% & 64.32\% \\

& Codex CLI
& 157 & 208 & 243
& 75.48\% & 64.61\% & 69.62\% \\

& OpenHands CLI
& 104 & 181 & 243
& 57.46\% & 42.80\% & 49.06\% \\
\midrule

\multirow{4}{*}{wang-2025}
& MetaRigor
& 328 & 375 & 486
& 87.47\% & 67.49\% & 76.19\% \\

& meta-pipe
& 90 & 152 & 486
& 59.21\% & 18.52\% & 28.21\% \\

& Codex CLI
& 350 & 457 & 486
& 76.59\% & 72.02\% & 74.23\% \\

& OpenHands CLI
& 256 & 397 & 486
& 64.48\% & 52.67\% & 57.98\% \\
\midrule

\multirow{4}{*}{qazi-2026}
& MetaRigor
& 200 & 239 & 252
& 83.68\% & 79.37\% & 81.47\% \\

& meta-pipe
& 94 & 147 & 252
& 63.95\% & 37.30\% & 47.12\% \\

& Codex CLI
& 162 & 227 & 252
& 71.37\% & 64.29\% & 67.64\% \\

& OpenHands CLI
& 153 & 234 & 252
& 65.38\% & 60.71\% & 62.96\% \\

\bottomrule
\end{tabularx}

\endgroup
\end{table}


\begin{table}[htbp]
\centering
\caption[Overall results across seven cases, with and without manuscript writing.]{%
\protect\hl{Overall results across seven cases, with and without manuscript writing.}%
}
\label{tab:g3}

\begingroup
\small
\setlength{\tabcolsep}{3pt}
\renewcommand{\arraystretch}{1.15}

\begin{tabularx}{\linewidth}{
  @{}ll*{6}{>{\centering\arraybackslash}X}@{}
}
\toprule
\textbf{Scope}
& \textbf{System}
& \textbf{Correct}
& \textbf{Expressed}
& \textbf{Reference}
& \textbf{Precision}
& \textbf{Recall}
& \textbf{F\textsubscript{1}} \\
\midrule

\multirow{4}{*}{\shortstack[l]{DE + RoB\\+ EC}}
& MetaRigor
& 1315 & 1623 & 1889
& 81.02\% & 69.61\% & 74.89\% \\

& meta-pipe
& 641 & 1011 & 1889
& 63.40\% & 33.93\% & 44.21\% \\

& Codex CLI
& 1200 & 1547 & 1889
& 77.57\% & 63.53\% & 69.85\% \\

& OpenHands CLI
& 871 & 1426 & 1889
& 61.08\% & 46.11\% & 52.55\% \\
\midrule

\multirow{4}{*}{\shortstack[l]{DE + RoB\\+ EC + MS}}
& MetaRigor
& 1539 & 1864 & 2133
& 82.56\% & 72.15\% & 77.01\% \\

& meta-pipe
& 814 & 1235 & 2133
& 65.91\% & 38.16\% & 48.34\% \\

& Codex CLI
& 1310 & 1764 & 2133
& 74.26\% & 61.42\% & 67.23\% \\

& OpenHands CLI
& 970 & 1610 & 2133
& 60.25\% & 45.48\% & 51.83\% \\

\bottomrule
\end{tabularx}

\endgroup
\end{table}

\subsection{Propagation of Result Identity Errors and Missingness}

SELECT2 was a randomized controlled trial comparing endovascular
thrombectomy plus medical therapy with medical therapy alone.
Its original report presented both mortality during hospitalization
and all cause mortality at 90 days.
The corresponding event counts were 42/178 versus 44/174 for the
former and 68/177 versus 71/171 for the latter
\citep{sarraj2023trial}.
The prespecified target was all cause mortality at 90 days.
MetaRigor selected the corresponding outcome, whereas OpenHands CLI
assigned the values for mortality during hospitalization to that
target (Table~\ref{tab:g4}).

\begin{table}[htbp]
\centering
\caption[Extracted results and downstream outcome binding for the SELECT2 target of all cause mortality at 90 days.]{%
\protect\hl{Extracted results and downstream outcome binding for the SELECT2 target of all cause mortality at 90 days.}%
}
\label{tab:g4}

\begingroup
\small
\setlength{\tabcolsep}{4pt}
\renewcommand{\arraystretch}{1.20}

\begin{tabularx}{\linewidth}{
  @{}
  >{\raggedright\arraybackslash}p{0.30\linewidth}
  *{3}{>{\centering\arraybackslash}X}
  @{}
}
\toprule

\textbf{Verification Item}
& \textbf{MetaRigor}
& \textbf{Codex CLI}
& \textbf{OpenHands CLI} \\
\midrule

Delivered values, thrombectomy group / medical therapy group
& 68/177 vs.\ 71/171
& Not reported
& 42/178 vs.\ 44/174
\\[5pt]

Actual outcome represented by the values
& All cause mortality at 90 days
& Not reported
& Mortality during hospitalization
\\[5pt]

Relationship to the prespecified target
& Outcome and time point matched
& Target result missing
& Outcome and time point mismatched
\\[5pt]

RoB status for this SELECT2 target
& HIGH
& NOT\_ASSESSED
& HIGH
\\

\bottomrule
\end{tabularx}

\endgroup
\end{table}

OpenHands CLI's final supplementary table continued to report
42/178 versus 44/174 under the target of mortality at 90 days.
This case illustrates how numerical values supported by the
original source can nevertheless become incorrect facts in the
final report when assigned to a target with a mismatched outcome
or follow up time.
Codex CLI did not deliver this target result, and its corresponding
RoB status was NOT\_ASSESSED.
Both MetaRigor and OpenHands CLI delivered a HIGH judgment.

Missing results can also propagate through the workflow.
For the outcome of mRS 0 or 1 at 90 days, Codex CLI delivered only
one of the six reference studies, leaving the remaining five
unevaluated in the downstream RoB stage.
The number of contributing studies was therefore insufficient
for statistical synthesis, and EC did not produce a complete
overall certainty downgrade judgment.

meta-pipe illustrates the downstream inheritance of numerical errors.
For the outcome of requiring constant care or death at 90 days,
defined as mRS scores of 5 or 6, in RESCUE-Japan LIMIT, the reference
number of events in the intervention group was 36, whereas meta-pipe
reported 58.
This value was subsequently carried forward into statistical
calculations and the inputs for manuscript writing.
These examples illustrate distinct pathways through which errors
propagate across stages, including result identity mismatches,
missing results, and incorrect numerical values.

\section[Supplementary Comparison under the GPT-6.1 Terra Configuration]{%
\protect\hl{Supplementary Comparison under the GPT-6.1 Terra Configuration}%
}
\label{app:terra}

We additionally conducted system comparisons under the GPT-6.1 Terra
configuration for DE, RoB, and EC.
The results are presented in Table~\ref{tab:terra_comparison}.
Each stage uses its corresponding inputs, evaluation targets,
and scoring criteria.
These results are reported separately from
Table~\ref{tab:comparative_performance} in the main text.
The EC implementation of meta-pipe is deterministic and does not
involve any Terra model calls.

\begin{table}[htbp]
\centering
\caption{
Supplementary comparative results under the GPT-6.1 Terra configuration.
}
\label{tab:terra_comparison}

\begingroup
\small
\setlength{\tabcolsep}{3pt}
\renewcommand{\arraystretch}{1.15}


\begin{tabularx}{\linewidth}{
  @{}l*{4}{>{\centering\arraybackslash}X}@{}
}
\toprule
& \multicolumn{4}{c}{\textbf{Data Extraction (DE)}} \\
\cmidrule(lr){2-5}

\textbf{Method}
& \textbf{Recall}$\uparrow$
& \textbf{F\textsubscript{1}}$\uparrow$
& \textbf{Field Acc.}$\uparrow$
& \textbf{Precision}$\uparrow$ \\
\midrule

Codex
& 61.54\%
& 74.77\%
& 63.65\%
& 95.24\%
\\[3pt]

OpenHands
& 58.97\%
& 72.33\%
& 59.78\%
& 93.50\%
\\[3pt]

meta-pipe
& 29.23\%
& 45.06\%
& 31.52\%
& 98.28\%
\\[3pt]

Manalyzer
& 29.74\%
& 45.31\%
& 33.33\%
& 95.08\%
\\[3pt]

\textbf{MetaRigor}
& 68.72\%
& 79.06\%
& 71.74\%
& 93.06\%
\\

\bottomrule
\end{tabularx}

\par\vspace{3pt}


\begin{tabularx}{\linewidth}{
  @{}l*{3}{>{\centering\arraybackslash}X}@{}
}
& \multicolumn{3}{c}{\textbf{Risk of Bias Assessment (RoB)}} \\
\cmidrule(lr){2-4}

\textbf{Method}
& \textbf{Domain}$\uparrow$
& \textbf{Judgment}$\uparrow$
& \textbf{High Risk}$\uparrow$ \\
\midrule

Codex
& 72.50\%
& 68.75\%
& 53.85\%
\\[3pt]

OpenHands
& 68.75\%
& 56.25\%
& 30.77\%
\\[3pt]

meta-pipe
& 65.00\%
& 62.50\%
& 38.46\%
\\[3pt]

Manalyzer
& N/A
& N/A
& N/A
\\[3pt]

\textbf{MetaRigor}
& 82.50\%
& 84.38\%
& 69.23\%
\\

\bottomrule
\end{tabularx}

\par\vspace{3pt}


\begin{tabularx}{\linewidth}{
  @{}l*{3}{>{\centering\arraybackslash}X}@{}
}
& \multicolumn{3}{c}{\textbf{Evidence Certainty (EC)}} \\
\cmidrule(lr){2-4}

\textbf{Method}
& \textbf{Domain}$\uparrow$
& \textbf{Downgrade}$\uparrow$
& \textbf{MAE}$\downarrow$ \\
\midrule

Codex
& 68.40\%
& 50.00\%
& 0.540
\\[3pt]

OpenHands
& 68.80\%
& 51.00\%
& 0.551
\\[3pt]

meta-pipe
& 57.80\%
& 31.00\%
& 0.990
\\[3pt]

Manalyzer
& N/A
& N/A
& N/A
\\[3pt]

\textbf{MetaRigor}
& 81.60\%
& 70.00\%
& 0.306
\\

\bottomrule
\end{tabularx}

\endgroup
\end{table}

Under this configuration, MetaRigor achieves a DE $F_1$ score
of 79.06\%, a recall of 68.72\%, and a field accuracy of 71.74\%.
For RoB, domain agreement, overall judgment agreement,
and high risk recall reach 82.50\%, 84.38\%, and 69.23\%,
respectively.
For EC, domain agreement and overall downgrade count agreement
reach 81.60\% and 70.00\%, respectively, with an MAE of 0.306.
Its advantages are most evident in result coverage,
field accuracy, and methodological assessment.

\end{document}

%% file: math_commands.tex
\usepackage{amsmath,amsfonts,bm}

\def\eqref#1{equation~\ref{#1}}

\def\1{\bm{1}}

\DeclareMathAlphabet{\mathsfit}{\encodingdefault}{\sfdefault}{m}{sl}
\SetMathAlphabet{\mathsfit}{bold}{\encodingdefault}{\sfdefault}{bx}{n}



%% file: iclr2027_conference.bib
@article{xu2025manalyzer,
  title={Manalyzer: End-to-end Automated Meta-analysis with Multi-agent System},
  author={Xu, Wanghan and Zhang, Wenlong and Ling, Fenghua and Fei, Ben and Hu, Yusong and Ma, Runmin and Zhang, Bo and Ren, Fangxuan and Lin, Jintai and Ouyang, Wanli and others},
  journal={arXiv preprint arXiv:2505.20310},
  year={2025}
}

@article{wang2025accelerating,
  title={Accelerating clinical evidence synthesis with large language models},
  author={Wang, Zifeng and Cao, Lang and Danek, Benjamin and Jin, Qiao and Lu, Zhiyong and Sun, Jimeng},
  journal={NPJ Digital Medicine},
  volume={8},
  number={1},
  pages={509},
  year={2025},
  publisher={Nature Publishing Group UK London}
}

@article{taherinezhad2026autosynthesis,
  title={AutoSynthesis: An agentic system for automated meta-analysis},
  author={Taherinezhad, Moein and Maier, Sebastian and Vitagliano, Gerardo and Pierri, Francesco and Feuerriegel, Stefan},
  journal={arXiv preprint arXiv:2607.15247},
  year={2026}
}

@inproceedings{hevia2025roboto2,
  title={ROBOTO2: An Interactive System and Dataset for LLM-assisted Clinical Trial Risk of Bias Assessment},
  author={Hevia, Anthony and Chintalapati, Sanjana and Lai, Veronica Ka Wai and Tam, Nguyen Thanh and Wong, Wai-Tat and Klassen, Terry P and Wang, Lucy Lu},
  booktitle={Proceedings of the 2025 Conference on Empirical Methods in Natural Language Processing: System Demonstrations},
  pages={12--25},
  year={2025}
}

@article{sterne2019rob,
  title={RoB 2: a revised tool for assessing risk of bias in randomised trials},
  author={Sterne, Jonathan AC and Savovi{\'c}, Jelena and Page, Matthew J and Elbers, Roy G and Blencowe, Natalie S and Boutron, Isabelle and Cates, Christopher J and Cheng, Hung-Yuan and Corbett, Mark S and Eldridge, Sandra M and others},
  journal={bmj},
  volume={366},
  year={2019},
  publisher={British Medical Journal Publishing Group}
}

@article{guyatt2025core,
  title={Core GRADE 2: choosing the target of certainty rating and assessing imprecision},
  author={Guyatt, Gordon and Zeng, Linan and Brignardello-Petersen, Romina and Prasad, Manya and De Beer, Hans and Murad, M Hassan and Iorio, Alfonso and Agarwal, Arnav and Yao, Liang and Agoritsas, Thomas and others},
  journal={Bmj},
  volume={389},
  year={2025},
  publisher={British Medical Journal Publishing Group}
}

@article{page2021prisma,
  title={The PRISMA 2020 statement: an updated guideline for reporting systematic reviews},
  author={Page, Matthew J and McKenzie, Joanne E and Bossuyt, Patrick M and Boutron, Isabelle and Hoffmann, Tammy C and Mulrow, Cynthia D and Shamseer, Larissa and Tetzlaff, Jennifer M and Akl, Elie A and Brennan, Sue E and others},
  journal={bmj},
  volume={372},
  year={2021},
  publisher={British Medical Journal Publishing Group}
}

@article{ahuja2026evisearch,
  title={{EviSearch}: A Human in the Loop System for Extracting and Auditing Clinical Evidence for Systematic Reviews},
  author={Ahuja, Naman and Mulla, Saniya and Khan, Muhammad Ali and Riaz, Zaryab Bin and Khakwani, Kaneez Zahra Rubab and Sonbol, Mohamad Bassam and Riaz, Irbaz Bin and Gupta, Vivek},
  journal={arXiv preprint arXiv:2604.14165},
  year={2026},
  note={Version 1},
  url={https://arxiv.org/abs/2604.14165v1}
}

@inproceedings{wang2025openhands,
  title={Openhands: An open platform for ai software developers as generalist agents},
  author={Wang, Xingyao and Li, Boxuan and Song, Yufan and Xu, Frank F and Tang, Xiangru and Zhuge, Mingchen and Pan, Jiayi and Song, Yueqi and Li, Bowen and Singh, Jaskirat and others},
  booktitle={International Conference on Learning Representations},
  volume={2025},
  pages={65882--65919},
  year={2025}
}

@article{van2021open,
  title={An open source machine learning framework for efficient and transparent systematic reviews},
  author={Van De Schoot, Rens and De Bruin, Jonathan and Schram, Raoul and Zahedi, Parisa and De Boer, Jan and Weijdema, Felix and Kramer, Bianca and Huijts, Martijn and Hoogerwerf, Maarten and Ferdinands, Gerbrich and others},
  journal={Nature machine intelligence},
  volume={3},
  number={2},
  pages={125--133},
  year={2021},
  publisher={Nature Publishing Group UK London}
}

@article{marshall2020trialstreamer,
  title={Trialstreamer: A living, automatically updated database of clinical trial reports},
  author={Marshall, Iain J and Nye, Benjamin and Kuiper, Jo{\"e}l and Noel-Storr, Anna and Marshall, Rachel and Maclean, Rory and Soboczenski, Frank and Nenkova, Ani and Thomas, James and Wallace, Byron C},
  journal={Journal of the American Medical Informatics Association},
  volume={27},
  number={12},
  pages={1903--1912},
  year={2020},
  publisher={Oxford University Press}
}

@misc{higgins2024cochrane,
  title     = {Cochrane Handbook for Systematic Reviews of Interventions, Version 6.5},
  author    = {Higgins, Julian P. T. and Thomas, James and Chandler, Jacqueline
               and Cumpston, Miranda and Li, Tianjing and Page, Matthew J.
               and Welch, Vivian A.},
  year      = {2024},
  note      = {Updated August 2024},
  publisher = {Cochrane},
  url       = {https://www.cochrane.org/handbook}
}

@incollection{li2024collecting,
  title     = {Chapter 5: Collecting Data},
  author    = {Li, Tianjing and Higgins, Julian P. T. and Deeks, Jonathan J.},
  booktitle = {Cochrane Handbook for Systematic Reviews of Interventions},
  editor    = {Higgins, Julian P. T. and Thomas, James and Chandler, Jacqueline
               and Cumpston, Miranda and Li, Tianjing and Page, Matthew J.
               and Welch, Vivian A.},
  edition   = {Version 6.5},
  publisher = {Cochrane},
  year      = {2024},
  note      = {Chapter last updated October 2019},
  url       = {https://training.cochrane.org/handbook/current/chapter-05}
}

@incollection{mckenzie2024defining,
  title     = {Chapter 3: Defining the Criteria for Including Studies and How They Will Be Grouped for the Synthesis},
  author    = {McKenzie, Joanne E. and Brennan, Sue E. and Ryan, Rebecca E.
               and Thomson, Hilary J. and Johnston, Renea V. and Thomas, James},
  booktitle = {Cochrane Handbook for Systematic Reviews of Interventions},
  editor    = {Higgins, Julian P. T. and Thomas, James and Chandler, Jacqueline
               and Cumpston, Miranda and Li, Tianjing and Page, Matthew J.
               and Welch, Vivian A.},
  edition   = {Version 6.5},
  publisher = {Cochrane},
  year      = {2024},
  note      = {Chapter last updated August 2023},
  url       = {https://training.cochrane.org/handbook/current/chapter-03}
}

@article{yao2022react,
  title={React: Synergizing reasoning and acting in language models},
  author={Yao, Shunyu and Zhao, Jeffrey and Yu, Dian and Du, Nan and Shafran, Izhak and Narasimhan, Karthik and Cao, Yuan},
  journal={arXiv preprint arXiv:2210.03629},
  year={2022}
}

@article{wang2025foundation,
  title={A foundation model for human-AI collaboration in medical literature mining},
  author={Wang, Zifeng and Cao, Lang and Jin, Qiao and Chan, Joey and Wan, Nicholas and Afzali, Behdad and Cho, Hyun-Jin and Choi, Chang-In and Emamverdi, Mehdi and Gill, Manjot K and others},
  journal={Nature communications},
  volume={16},
  number={1},
  pages={8361},
  year={2025},
  publisher={Nature Publishing Group UK London}
}

@article{sarraj2023trial,
  title={Trial of endovascular thrombectomy for large ischemic strokes},
  author={Sarraj, Amrou and Hassan, Ameer E and Abraham, Michael G and Ortega-Gutierrez, Santiago and Kasner, Scott E and Hussain, M Shazam and Chen, Michael and Blackburn, Spiros and Sitton, Clark W and Churilov, Leonid and others},
  journal={New England Journal of Medicine},
  volume={388},
  number={14},
  pages={1259--1271},
  year={2023},
  publisher={Mass Medical Soc}
}

@article{lin2026meta,
  title={meta-pipe: An LLM-agent pipeline for end-to-end automated systematic review and meta-analysis},
  author={Lin, Hsieh-Ting and Yeh, Jiunn-Tyng},
  journal={arXiv preprint arXiv:2606.28363},
  year={2026}
}

@article{marshall2016robotreviewer,
  title={{RobotReviewer}: evaluation of a system for automatically assessing bias in clinical trials},
  author={Marshall, Iain J. and Kuiper, Jo{\"e}l and Wallace, Byron C.},
  journal={Journal of the American Medical Informatics Association},
  volume={23},
  number={1},
  pages={193--201},
  year={2016},
  doi={10.1093/jamia/ocv044}
}
